\documentclass[preprint]{elsarticle}

\usepackage{lineno,hyperref}
\modulolinenumbers[5]

\journal{Journal of \LaTeX\ Templates}

\usepackage{framed,multirow}
\usepackage{amsmath}
\usepackage{blkarray}
\usepackage{placeins}
\usepackage{amssymb, bm}
\usepackage{latexsym}
\usepackage[ruled,vlined, algo2e]{algorithm2e}
\usepackage[utf8x]{inputenc}
\setcitestyle{square}
\usepackage{multicol,setspace}

\usepackage{graphicx}
\usepackage{array}
\newcolumntype{P}[1]{>{\centering\arraybackslash}p{#1}}
\begin{document}

\begin{frontmatter}

\title{Speed-up and Multi-view Extensions to Subclass Discriminant Analysis}%\tnoteref{mytitlenote}}
%\tnotetext[mytitlenote]{Fully documented templates are available in the elsarticle package on \href{http://www.ctan.org/tex-archive/macros/latex/contrib/elsarticle}{CTAN}.}

%% or include affiliations in footnotes:
\author[mymainaddress]{Kateryna Chumachenko\corref{mycorrespondingauthor}}
\cortext[mycorrespondingauthor]{Corresponding author}
\ead{kateryna.chumachenko@tuni.fi}

\author[mymainaddress,mythirdaddress]{Jenni Raitoharju}
\ead{jenni.raitoharju@tuni.fi}

\author[mysecondaryaddress]{Alexandros Iosifidis}
\ead{alexandros.iosifidis@eng.au.dk}

\author[mymainaddress]{Moncef Gabbouj}
\ead{moncef.gabbouj@tuni.fi}

\address[mymainaddress]{Faculty of Information Technology and Communication Sciences, Tampere University, FI-33720 Tampere, Finland}
\address[mysecondaryaddress]{Department of Engineering, Aarhus University, DK-8200 Aarhus, Denmark}
\address[mythirdaddress]{Programme for Environmental Information, Finnish Environment Institute, FI-40500 Jyväskylä, Finland}

\begin{abstract}
In this paper, we propose a speed-up approach for subclass discriminant analysis and formulate a novel efficient multi-view solution to it. The speed-up approach is developed based on graph embedding and spectral regression approaches that involve eigendecomposition of the corresponding Laplacian matrix and regression to its eigenvectors. We show that by exploiting the structure of the between-class Laplacian matrix, the eigendecomposition step can be substituted with a much faster process. Furthermore, we formulate a novel criterion for multi-view subclass discriminant analysis and show that an efficient solution to it can be obtained in a similar manner to the single-view case. We evaluate the proposed methods on nine single-view and nine multi-view datasets and compare them with related existing approaches. Experimental results show that the proposed solutions achieve competitive performance, often outperforming the existing methods. At the same time, they significantly decrease the training time.
\end{abstract}


\end{frontmatter}

%\linenumbers

\section{Introduction}
\label{sec1}
In the modern world, large amounts of data available for training of machine learning algorithms result in their applicability and efficiency in different subject areas \cite{scalecskda, mvdareid}. However, when the dimensionality of data is high, the algorithms can become susceptible to the well-known curse of dimensionality, stating that in the cases of high-dimensional data, its representation becomes sparse and, therefore, huge amounts of training data are required for the estimation of the parameters of a machine learning method. To address this problem, many dimensionality reduction methods were proposed over the recent years, acquiring an important role within the machine learning field. The objective of the dimensionality reduction methods is to determine a feature space, projection onto which results in a lower dimensionality of data, while preserving properties of the data that are of interest for the problem at hand. 

Subspace learning methods can be divided into unsupervised and supervised ones, i.e., those relying solely on the structure of data and those exploiting additional class label information provided by experts. Among the unsupervised dimensionality reduction methods, probably the most common one is Principal Component Analysis (PCA) \cite{pca}, that projects the data onto the subspace where the data has the highest variance. 

Supervised subspace learning methods assume that during training the data is given with class labels. Therefore, they lead to enhanced class discrimination compared to unsupervised methods and they are more suitable for classification problems. One of the most well-known methods incorporating the information on class distribution is Linear Discriminant Analysis (LDA) \cite{lda, fisherwang}, where the optimal subspace is obtained by optimizing the Fisher - Rao's criterion \cite{fischerrao} that is defined over the within-class and between-class scatter matrices, under the assumption that the classes are unimodal and follow normal distribution. While incorporating the class label information, LDA can only define a subspace of at most $d$ dimensions, where $d$ is the rank of the between-class scatter matrix, which is equal to $C$-$1$ for the case of $C$ classes. 

The assumption of the class unimodality in LDA limits its performance in problems where classes form subclasses, i.e., classes are represented by multiple disjoint distributions. In order to address this limitation, approaches incorporating the subclass information in the optimization problem solved for determining the discriminant subspace have been proposed. Methods following this approach are the Subclass Discriminant Analysis (SDA) \cite{sda}, Clustering Discriminant Analysis (CDA) \cite{cda}, and Subclass Marginal Fisher Analysis (SMFA) \cite{sge}. In addition to better describing the classes' distributions, these methods are also able to determine discriminant subspaces of higher dimensionalities, since the maximum dimensionality of the learned feature space is limited by the rank of a modified between-class scatter matrix which is bounded by the total number of subclasses.

One of the main drawbacks of the subspace learning methods lies in the low speed for high-dimensional data and large datasets. For speeding up the training process several approaches have been proposed, including approximate solutions \cite{scalecskda}, incremental learning \cite{kwak}, and speed-up solutions \cite{krda, cskda, iosifidis2015kernel, fastfda, iosifidis2017approximate}. However, none of these approaches are able to overcome the assumption of unimodality and limitations related to limited dimensionality of the learnt subspace. Therefore, we aim to bridge the speed-up and multi-modal solutions and propose a method that can overcome all of the main limitations of LDA and related methods at once. We achieve this by proposing a speed-up approach for Subclass Discriminant Analysis that already allows to overcome the unimodality assumption and limited potential dimensionality of the learned subspace. In this paper, we propose a speed-up approach for SDA and its kernelized form, i.e., Kernel Subclass Discriminant Analysis (KSDA) \cite{ksda}. The proposed approach is based on graph embedding \cite{sge, graphsub} and exploitation of the structure of the between-class Laplacian matrix. 

In some problems, the descriptions of the same items from multiple differently distributed modalities might be available, resulting in multiple modalities of the data. Such problems are referred to as multi-view or multimodal problems. The nature of multi-view problems is similar to the way humans perceive the world and take decisions, as the real-world data is not limited to one source, but consists of, e.g., visual and audio signals, tactile sensations. The data from different modalities is perceived by human and the decision is made by combining information from different sources. A similar approach is followed by multi-view subspace learning methods, where the combination of the information coming from different views is performed by defining a latent feature space, jointly determined using data from all available views during the training process. Moreover, the views can have different dimensionalities. An example of a multi-view problem is the classification of video sequences using their two views, i.e., audio and visual signals. 

Extensions of supervised subspace learning methods to the multi-view case include the Multi-view Discriminant Analysis (MDA) \cite{mvda} that defines a variant of the LDA criterion to incorporate information from multiple views. In \cite{mvda}, the between-class scatter is maximized regardless of the difference between inter-view and intra-view covariances, while the within-class scatter is minimized. 
Multi-view Common Component Discriminant Analysis proposes a way to address the nonlinearity, view discrepancy and discriminability jointly by incorporating both label information and geometric information during subspace learning \cite{mvccda}. In order to address the problem of multi-label classification with a high number of classes on a multi-view dataset, a Multi-view Label Embedding model was proposed \cite{mvml}. Besides, for the problems with incomplete or incompletely labeled multi-view data, a unified subspace learning framework has been proposed \cite{incompletesl}.
In addition, several multi-view extensions of LDA have been recently proposed, including Standard Multi-view Discriminant Analysis (SMvDA) and Multi-view Modular Discriminant Analysis (MvMDA) \cite{mvLDA}. Being extensions of LDA, these methods have similar limitations: the assumption of the unimodality of data within each view and maximal number of dimensions bounded by the number of classes. In this work, we propose an approach to overcome these limitations by introducing Multi-view Subclass Discriminant Analysis, as well as its kernelized form, and show that the solution for its optimization problem can be obtained by following a fast and efficient process.
%In order to maximize intra-view discriminant information, Generalized Multi-view Analysis (GMA) was proposed \cite{gma}. In order to simultaneously consider the cross-modal correlation and intra-modal discriminant information the Cross-Modal Deep Discriminant Analysis was recently proposed \cite{crossmodaldeep}, and Multi-view Local Discrimination, and Canonical Correlation Analysis were proposed in order to incorporate the inter-modal discriminant information \cite{multiviewlocal}. In addition to that, several multi-view extensions of LDA have been recently proposed, including Standard Multi-view Discriminant Analysis (SMvDA) and Multi-view Modular Discriminant Analysis(MvMDA) \cite{mvLDA}. Being extensions of LDA, these methods have similar limitations: the assumption of the unimodality of data within each view and maximal number of dimensions bounded by the number of classes. In this work, we propose an approach to overcome these limitations by introducing Multi-view Subclass Discriminant Analysis, as well as its kernelized form, and show that the solution for its optimization problem can be obtained by following a fast and efficient process.

The proposed work brings the following contributions:
\begin{itemize}
    \item First, we define the Graph Embedding and Spectral Regression -based formulations of Subclass Discriminant Analysis and Kernel Subclass Discriminant Analysis.
    \item Second, based on the previously-defined formulations, we show how the exploitation of the properties of constant-sum block matrices and the specific structure of the between-class Laplacian matrix can be utilized for speeding up the method. The speed up is achieved by solving the slow eigendecomposition step, which is the main computational bottleneck of SDA, using a fast process of creation of target vectors based on the above-mentioned properties.
    \item Third, we define a novel multi-view formulation of SDA and KSDA that allows to apply the methods on multi-view data and take into account the potential multi-modalities present in different views. We show how the speed-up approach defined for the single-view formulation can be modified for the multi-view case, resulting in a speed up of the algorithm.
\end{itemize}

\section{Related Work}
This section describes the previous works related to the proposed supervised subspace learning methods. 

Let us consider a set of $N$ ${D}$-dimensional vectors $\mathbf{X} =[\mathbf{x}_1, \mathbf{x}_2, ..., \mathbf{x}_N] \in \mathbb{R}^{D}$, each belonging to a class indicated by the corresponding label $c_i$. We define the subspace learning problem as searching for the $d$-dimensional feature space, with $d < D$, that provides the highest class separability of the data in $\mathbf{X}$ when projected onto that space. Most dimensionality reduction methods, including LDA, SDA, CDA, and SMFA optimize the Fisher-Rao's criterion \cite{fischerrao}:
\begin{equation}
    \mathcal{J}(\mathbf{W}) = {\underset{\mathbf{W}^T\mathbf{W} = \mathbf{I}}{argmin}}\:\frac{Tr(\mathbf{W}^T\mathbf{S}_w\mathbf{W})}{Tr(\mathbf{W}^T\mathbf{S}_b\mathbf{W})},
\end{equation} where $Tr()$ denotes the trace operator, $\mathbf{S}_w$ and $\mathbf{S}_b$ are symmetric positive semi-definite matrices, referred to as within-class and between-class scatter matrices. The main differences between the subspace learning methods lie in the definition of these matrices. 
LDA \cite{lda} assumes that each class is unimodal and seeks to find a space, projection onto which would result in compact classes lying far from each other, hence, resulting in high discrimination between classes. The within-class and between-class scatter matrices are defined as \begin{equation} \mathbf{S}_w = \sum_{i=1}^{C}\sum_{j=1}^{N_i}(\mathbf{x}_{ij} - \bm{\mu}_i)(\mathbf{x}_{ij}-\bm{\mu}_i)^T,\end{equation}
\begin{equation} \mathbf{S_b} = \sum_{i=1}^{C}(\bm{\mu}_{i} - \bm{\mu})(\bm{\mu}_{i}-\bm{\mu})^T,\end{equation}
where $C$ is the number of classes, $\mathbf{\mu}$ is the mean of data, $\mathbf{\mu}_i$ is the mean of class $i$, $N_i$ is the number of samples in class $i$ and $\mathbf{x}_{ij}$ is the $j^{th}$ sample of class $i$.

Many extensions to LDA have been proposed over the recent  years. Methods relaxing the assumption of LDA about normally distributed classes and the limitations on the dimensionality of the learned subspace in binary problems have been recently proposed in \cite{rvda,krefda,csrefda}. %Stochastic Discriminant Analysis was proposed as an extension to LDA for problems where low-dimensional projections of datasets having several classes are needed. Similarities between samples are represented by transforming distances to joint probabilities using a transformation resembling Student’s t-distribution \cite{stochasticda}.

CDA \cite{cda} relaxes the assumption on unimodal classes and applies clustering techniques to incorporate the subclass structure of the data in the training process.
SMFA relies on a framework of Subclass Graph Embeddings \cite{sge}, where the dimensionality reduction problem is described from a graph embedding perspective. The problem is defined by intrinsic and penalty graph matrices, which are built relying on the label information of $k$ nearest neighbors of the data points, as defined by the Euclidean distance or some other distance metric. The intrinsic graph matrix represents the compactness within the subclass, while the penalty graph matrix enforces penalization to ensure inter-class separability. 

%Further in this section we describe SDA, KSDA, Spectral Regression Discriminant Analysis, Kernel Regression and Approximate Kernel Regression in more detail.
\subsection{Graph Embedding Framework}
A framework that considers different subspace learning algorithms from a graph embedding perspective has been proposed in \cite{sgefram}. A further extension to subclass-based methods has been proposed in \cite{sge}. In both of these frameworks, data is described using two undirected weighted graphs: an intrinsic graph $\mathbf{G} = \{\mathbf{X},\bm{\Omega}\}$ with vertex set $\mathbf{X}$ and similarity matrix $\bm{\Omega}$, and a penalty graph $\mathbf{G}^p = \{\mathbf{X},\bm{\Omega}^p\}$ that represents the similarity characteristics of the data that are desired to be suppressed in the learned space. Each vertex in $\mathbf{X}$ corresponds to a data sample. For each pair of vertices in $\mathbf{X}$, $\bm{\Omega}$ measures their similarity by means of some similarity criterion, e.g., Gaussian similarity. Then, the diagonal degree matrix $\mathbf{D}$, the Laplacian matrix $\mathbf{L}$ are defined as 

\begin{equation}
\centering
\begin{aligned}
\mathbf{L} = \mathbf{D} - \bm{\Omega},\\
\mathbf{D}^{ii} = \sum_{j\ne i}\bm{\Omega}^{ij},
\end{aligned}
\label{dp}
\end{equation}
i.e., the degree matrix $\mathbf{D}$ at position $(i,i)$ has the value of the sum of all values of $\bm{\Omega}$ across $i$'$th$ row or column, as $\bm{\Omega}$ is symmetric. The penalty Laplacian matrix $\mathbf{L}_b$ can be defined similarly using the penalty matrix $\bm{\Omega}_p$. The goal of graph embedding is to find such a low-dimensional representation relationship among the vertices in $\mathbf{X}$ that incorporates the similarity relationship outlined in $\mathbf{G}$ in the best way. 

Let us define a low-dimensional representation of vertices in $\mathbf{X}$ as $\mathbf{y} = [\mathbf{y}_1, ..., \mathbf{y}_i]$. Then, the objective function that solves the problem of finding such a projection matrix $\mathbf{W}$ that would incorporate the similarity between the vertices in $\mathbf{X}$ can be defined as follows:
\begin{equation}
    \mathbf{y}^* ={argmin}\:\sum_{i\ne j}||\mathbf{y}_i - \mathbf{y}_j||^2\bm{\Omega}_{ij} = {argmin}\: \mathbf{y}^T\mathbf{Ly} = {argmin}\:\mathbf{w}^T\mathbf{XLX}^T\mathbf{w}, \label{geobj}
\end{equation}
given $\mathbf{y} = \mathbf{X}^T \mathbf{w}$ is the projection of the data point to a subspace. Similarly, a maximization problem can be defined using $\mathbf{L}_b$. Such formulation allows to reformulate various subspace learning methods and take advantage of the new formulations, as will be shown further.

\subsection{Subclass Discriminant Analysis}
In order to relax the class unimodality assumption of LDA, SDA \cite{sda} expresses each class by a set of subclasses that are obtained by applying clustering on the class data. The difference between CDA and SDA lies in the definition of the within-class and between-class scatter matrices. In SDA, the total scatter matrix $\mathbf{S}_t$ is minimized instead of the within-class scatter as $\mathbf{S}_t = \mathbf{S}_b + \mathbf{S}_w$. SDA uses the following definitions: 
\begin{equation}
\mathbf{S}_t = \sum_{q=1}^{N}(\mathbf{x}_q - \bm{\mu})(\mathbf{x}_q - \bm{\mu})^T,
\end{equation}
\begin{equation}
\mathbf{S}_b = \sum_{i=1}^{C-1}\sum_{l=i+1}^{C}\sum_{j=1}^{d_i}\sum_{h=1}^{d_l}p_{ij}p_{lh}(\bm{\mu}_{ij} - \bm{\mu}_{lh})(\bm{\mu}_{ij} - \bm{\mu}_{lh})^T,
\end{equation}
where $\bm{\mu}$ is the mean of data, $i$ and $l$ are class labels, $j$ and $h$ are subclass labels, $p_{ij}$ and $p_{lh}$ are the subclass priors, $p_{ij} = \frac{N_{ij}}{N}$, where $N_{ij}$ is the number of samples in subclass $j$ of class $i$ and $N$ is the total number of samples in $\mathbf{X}$.
The solution of (1) is given by solving the generalized eigendecomposition problem
\begin{equation}\mathbf{S}_t\mathbf{w} = \lambda \mathbf{S}_b\mathbf{w}.\end{equation}
The obtained eigenvectors $[\mathbf{w}_1, \mathbf{w}_2, ..., \mathbf{w}_d]$ that correspond to $d$ minimal eigenvalues form a projection matrix $\mathbf{W}$. The projected data point $\mathbf{y}_i$ can be computed as $\mathbf{y}_i = \mathbf{W}^T\mathbf{x}_i$. 

It is trivial to see that for the data centered at $\bm{\mu}$, $\mathbf{S}_t = \mathbf{XX}^T$. In addition, the representation of $\mathbf{S}_b$ can be defined using the Graph Embedding framework as follows:
\begin{equation}\mathbf{S}_b = \mathbf{XL}_b\mathbf{X}^T,\end{equation}
\begin{equation}\mathbf{L}_b(i,j) = \begin{cases}
			\frac{N - N_{c_{i}}}{N^2 N_{ch}}, & \text{if $z_i = z_j = h$}\\
            0, & \text{if $z_i \ne z_j$, $c_i = c_j$} \\
            - \frac{1}{N^2}, & \text{if $c_i \ne c_j$}
		 \end{cases},\end{equation}
where $c_i$ is the class label of $\mathbf{x}_i$, and $z_i$ is the subclass label of $\mathbf{x}_i$, $N_c$ is the number of samples in class $c$ and $N_{ch}$ is the number of samples in subclass $h$ of class $c$.

The objective function of SDA can be reformulated into a maximization problem (11), and exploiting the formulations in (9) and (10), the solution is given by the generalized eigendecomposition problem (12), and the projection matrix is obtained by selecting the eigenvectors corresponding to maximal eigenvalues. 
\begin{equation}\mathcal{J}(\mathbf{W}) = {\underset{\mathbf{W}^T\mathbf{W} = \mathbf{I}}{argmax}}\:\frac{Tr(\mathbf{W}^T\mathbf{S}_b\mathbf{W})}{Tr(\mathbf{W}^T\mathbf{S}_t\mathbf{W})},\end{equation}
\begin{equation}\mathbf{L}_b\mathbf{X}^T\mathbf{v} = \lambda \mathbf{X}^T\mathbf{v}.\end{equation}

\subsection{Kernel Subclass Discriminant Analysis}
Kernel methods are widely used in machine learning to overcome the limitation of the linear separability, which is rarely present in real-world problems. In order to nonlinearly map each data point $\mathbf{x}_i$ from the space $\mathbb{R}^D$ to its image $\bm{\phi}_i$ in some space $\mathcal{F}$, the nonlinear function $\phi(\mathbf{x})$ is defined, i.e., $\phi(\mathbf{x}_i) \in \mathcal{F}$. The dimensionality of $\mathcal{F}$ depends on the choice of the function and can be arbitrary. A linear projection is then defined in $\mathcal{F}$, i.e. $\mathbf{y}_i = \mathbf{W}^T\phi(\mathbf{x}_i)$. 

The conventional approach to solving the nonlinear problems involves the exploitation of kernel function defined over a pair of data points in $\mathbf{X}$ that maps them to the dot product of their projections in $\mathcal{F}: k(\mathbf{x}_1, \mathbf{x}_2) = \phi(\mathbf{x}_1)^T\phi(\mathbf{x}_2)$ and formulating the problem accordingly. By exploiting the dot product representation, the explicit mapping of each data point $\mathbf{x}_i$ in $\mathbf{X}$ to its image $\bm{\phi}_i = \phi(\mathbf{X})$ can be omitted, hence, avoiding the issues related to the arbitrary dimensionality of $\mathcal{F}$. The $N \times N$ kernel matrix $\mathbf{K}$ is defined as $\mathbf{K}_{ij} = k(\mathbf{x}_i,\mathbf{x}_j)$. It is easy to note that since $k(\mathbf{x}_i, \mathbf{x}_j) = \phi(\mathbf{x}_i)^T\phi(\mathbf{x}_j), \mathbf{K} = \bm{\Phi}^T\bm{\Phi}$, where $\bm{\Phi} = [\phi(\mathbf{x}_1), \phi(\mathbf{x}_2), ..., \phi(\mathbf{x}_N)]$.
According to the Representer Theorem \cite{repr}, $\mathbf{W}$ can be represented as a linear combination of data in $\mathcal{F}$
\begin{equation}\mathbf{W} = \bm{\Phi}\mathbf{A}.\end{equation}
Therefore, $\mathbf{y}_i = \mathbf{W}^T\phi(\mathbf{x}_i) = \mathbf{A}^T\bm{\Phi}^T\phi(\mathbf{x}_i) = \mathbf{A}^T\mathbf{k}_i.$

The kernelization of the SDA can be easily obtained by exploiting the modified representation of $\mathbf{S}_b$ and $\mathbf{S}_t$ (9) \cite{ksda}. Here we can assume that data is centered in $\mathcal{F}$. The kernel matrix of the centered data can be obtained as in (14) \cite{kcenter}
\begin{equation}
\mathbf{K}^c = (\mathbf{I}- \mathbf{E}_N) \mathbf{K} ( \mathbf{I}-\mathbf{E}_N),
\end{equation}
\begin{equation}
\mathbf{E}_N = \frac{1}{N}\mathbf{1}_N \mathbf{1}_{N}^T,
\end{equation}
where $\mathbf{1}_N \in \mathbb{R}^{N}$is a vector of ones. 
%\begin{equation}\mathbf{K}_{ij}^c = \mathbf{K}_{ij} - \mathbf{k}_i\mathbf{1}_j^T - \mathbf{1}_i\mathbf{k}_j^T + q\mathbf{1}_i\mathbf{1}_j^T,\end{equation}
%\begin{equation}\mathbf{k}_i = \frac{1}{N}\sum_k\mathbf{K}_{ik},\end{equation} \begin{equation}q = \frac{1}{N^2}\sum_{ij}\mathbf{K}_{ij}.\end{equation}

After mean-centering $\phi(\mathbf{X})$, $\mathbf{S}_{kt}$ and $\mathbf{S}_{kb}$ are given as follows:
\begin{equation}\mathbf{S}_{kt} = \sum_{i=1}^{N}(\bm{\phi}_i - \bm{\bar{\phi}})(\bm{\phi}_i - \bm{\bar{\phi}})^T = \bm{\phi}\bm{\phi}^T,\end{equation}
\begin{equation}\begin{aligned}\mathbf{S}_{kb} = \sum_{i=1}^{C-1}\sum_{l=i+1}^{C}\sum_{j=1}^{d_i}\sum_{h=1}^{d_l}p_{ij}p_{lh}(\bm{\bar{\phi}}_{ij} - \bm{\bar{\phi}}_{lh})(\bm{\bar{\phi}}_{ij} - \bm{\bar{\phi}}_{lh})^T 
= \bm{\phi} \mathbf{L}_b\bm{\phi}^T,\end{aligned}\end{equation}
where $\bar{\bm{\phi}}_{ij}$ is the mean of the subclass $j$ of class $i$ in $\mathcal{F}$, $\mathbf{L}_b$ is the between-class Laplacian matrix defined in (10), and $\bar{\bm{\phi}}$ is the mean of the data in $\mathcal{F}$.
%The objective is the optimization of the following function:
%\begin{equation}J(v) = {\underset{W}{argmax}}\frac{Tr(W^TS_{kb}W)}{Tr(W^TS_{kw}W)}\end{equation}
Exploiting (13,17-18), the solution to KSDA is given by the generalized eigendecomposition problem 
%\begin{equation}S_{kb}v = \lambda S_{kw}v\end{equation}
%It can be seen that for the kernel case and the mean-centered data $S_{kw} = \Phi \Phi^T$ and $S_{kb} = \Phi J_b \Phi^T$, therefore the generalized eigendecomposition problem can be respresented as 
\begin{equation}\bm{\Phi} \mathbf{L}_b\bm{\Phi}^T\bm{\Phi}\mathbf{a} = \lambda \bm{\Phi}\bm{ \Phi}^T \bm{\Phi}\mathbf{a} => %\Phi^T\Phi L_b\Phi^T\Phi a = \lambda \Phi^T \Phi \Phi^T \Phi a,
\end{equation}
\begin{equation}
\mathbf{KL}_b\mathbf{Ka} = \lambda \mathbf{K K a} => \mathbf{L}_b\mathbf{Ka} = \lambda \mathbf{Ka}.\end{equation}
%\begin{equation}\Phi J_b\Phi^T\Phi a = \lambda \Phi \Phi^T \Phi a\end{equation}
%\begin{equation}\Phi^T\Phi J_b\Phi^T\Phi a = \lambda \Phi^T \Phi \Phi^T \Phi a\end{equation}
%\begin{equation}KJ_bKa = \lambda K K a\end{equation}
%\begin{equation}J_bKa = \lambda Ka\end{equation}

\subsection{Multi-view Extensions to Linear Discriminant Analysis}
In multi-view learning, the data $\mathbf{X} = diag(\mathbf{X}_1, \mathbf{X}_2, ..., \mathbf{X}_V)$ is described from $V$ views and we seek to find $V$ matrices $\mathbf{W}_v$ that project the data $\mathbf{X}_v$ from
all views $v = 1, ..., V$ to a common (latent) space,  where the separability between the classes is the highest.
A generalized framework for multi-view subspace learning, that includes many of the existing methods as special cases, was proposed in \cite{mvLDA}. Here, the optimization problem is defined as
\begin{equation}\mathcal{J}(\mathbf{W}) = {\underset{\mathbf{W}^T\mathbf{W} = \mathbf{I}}{argmax}}\:\frac{Tr(\mathbf{W}^T\mathbf{PW})}{Tr(\mathbf{W}^T\mathbf{QW})},\end{equation}
where $\mathbf{P}$ and $\mathbf{Q}$ are the inter-view and intra-view covariance matrices. The solution is obtained by solving the generalized eigendecomposition problem 
\begin{equation}\mathbf{PW} = \bm{\rho} \mathbf{QW},\end{equation}
\begin{equation}\mathbf{W} = \left(\begin{array}{c}
     \mathbf{W}_{1} \\
     \mathbf{W}_{2} \\
     ... \\
     \mathbf{W}_{V} \end{array}\right),\end{equation}
     where $\mathbf{W}_v$ is the projection matrix of the view $v$. The feature vectors in the latent space are obtained as $\mathbf{Y}_v = \mathbf{W}_v^T\mathbf{X}_v$, where $\mathbf{X}_v$ is data representation in the view $v$.
     \arraycolsep=1pt
     Here, 
     %\begin{equation}\mathbf{P} = \left(\begin{array}{cccc}
     %\mathbf{P}_{11} & \mathbf{P}_{12} & ... &\mathbf{P}_{1V}\\
     %\mathbf{P}_{21} & \mathbf{P}_{22} & ... &\mathbf{P}_{2V} \\
     %... \\
     %\mathbf{P}_{V1} & \mathbf{P}_{V2} & ... &\mathbf{P}_{VV} \end{array}\right),
     %\mathbf{Q} = \left(\begin{array}{cccc}
     %\mathbf{Q}_{11} & 0 & ... &0\\
     %0 & \mathbf{Q}_{22} & ... &0 \\
     %... \\
     %0 & 0 & ... &\mathbf{Q}_{VV}\end{array}\right),\end{equation}
     
%\begin{equation}\mathbf{P}_{ij} = \mathbf{X}_i\mathbf{L}_b\mathbf{X}_j^T,\end{equation}
%\begin{equation}\mathbf{Q}_{ii} = \mathbf{X}_i\mathbf{L}_w\mathbf{X}_i^T.\end{equation}
\begin{equation}\mathbf{P} = \mathbf{X}\mathbf{L}_b\mathbf{X}^T,\end{equation}
\begin{equation}\mathbf{Q} = \mathbf{X}\mathbf{L}_w\mathbf{X}^T,\end{equation}
\begin{equation}\mathbf{X} = \left(\begin{array}{cccc}
    \mathbf{X}_1 & 0 & ... & 0 \\
    0 & \mathbf{X}_2 & ... & 0 \\
    0 & 0 & ... & \mathbf{X}_V \end{array}\right),\end{equation}
    
    \begin{equation}\mathbf{L}_b = \left(\begin{array}{cccc}
     \mathbf{L}_{b11}  & \mathbf{L}_{b12} & ... & \mathbf{L}_{bV1}\\
     \mathbf{L}_{b12}  & \mathbf{L}_{b22} & ... & \mathbf{L}_{bV2} \\
     ... & ... & ... & ...  \\
     \mathbf{L}_{b1V}  & \mathbf{L}_{b2V} & ... & \mathbf{L}_{bVV}  \end{array}\right),\end{equation}
     \begin{equation}\mathbf{L}_w = \left(\begin{array}{cccc}
     \mathbf{L}_{w11}  & 0 & ... & 0\\
     0  & \mathbf{L}_{w22} & ... & 0 \\
     ... & ... & ... & ...  \\
     0  & 0 & ... & \mathbf{L}_{wvv}  \end{array}\right),\end{equation}
     where $\mathbf{L}_{bij}$ is either $\mathbf{L}_{bij}^*$ or $\mathbf{\hat{L}}_{bij}$, as defined below, $i$ and $j$ are the view labels, and $V$ is the number of views.

Using the above notations, SMvDA aims to maximize the distance between the class means regardless of the view and defines $\mathbf{L}_{bij}$ as
\begin{equation}\mathbf{L}_{bij}^* = \begin{cases}
			2\sum_{p=1}^C\sum_{\underset{q\ne p}{q=1}}^C(\frac{V}{N_p^2}\mathbf{e}_p\mathbf{e}_p^T - \frac{1}{N_pN_q}\mathbf{e}_p\mathbf{e}_q^T), & \text{if $i = j$}\\
            -2\sum_{p=1}^C\sum_{\underset{q\ne p}{q=1}}^C\frac{1}{N_pN_q}\mathbf{e}_p\mathbf{e}_q^T, & \text{if $i \ne j$} \\
            \end{cases},\end{equation} where $\mathbf{e}_p$ is $N$-dimensional class vector with 1s at the positions corresponding to the samples belonging to class $p$ and 0s elsewhere, $i$ and $j$ are views, and $C$ is the number of classes.

The MvMDA maximizes the distances between the centers of different classes across different views:
 \begin{equation}\mathbf{\hat{L}}_{bij} = 2\sum_{p=1}^C\sum_{q=1}^C(\frac{1}{N_p^2}\mathbf{e}_p\mathbf{e}_p^T - \frac{1}{N_pN_q}\mathbf{e}_p\mathbf{e}_q^T).\end{equation}
In both cases, the intra-view Laplacian matrix $\mathbf{L}_{w}$ is defined as in (27), where
%\begin{equation}\mathbf{Q}_{ii} \\ = \sum_{c=1}^C\sum_{k=1}^{N_c}(\mathbf{x}_k^i - \bm{\mu}_c^i)(\mathbf{x}_k^i - \bm{\mu}_c^i)^T  =  \mathbf{X}_i\mathbf{L}_{wii}\mathbf{X}_i^T,\end{equation}
\begin{equation}\mathbf{L}_{wii} = \mathbf{I} - \sum_{c=1}^C\frac{1}{N_c}\mathbf{e}_c\mathbf{e}_c^T,\end{equation}
where $i$ is the view label, $c$ is the class label, $C$ is the total number of classes, and $\mathbf{I}$ is the identity matrix.
Similarly, the solution to Kernel MvMDA and Kernel SMvDA is given by optimizing

%\begin{equation}J(A) = \underset{A_v^TKA_v = I,v=1,..,V}{argmax}\frac{Tr(\sum_{i=1}^V\sum_{j=1}^VA_i^TK_iL_b^*K_jA_j)}{Tr(\sum_{i=1}^VA_i^TK_iL_wK_iA_i)}\end{equation}
%where $L_b^*$ is $L_b$ or $\hat{L_b}$. The problem is solved by (15) and $Y_v = A_v^T\Phi_v^T\Phi_v = A_v^TK_v$.

\begin{equation} \mathcal{J}(\mathbf{A}) = {\underset{\mathbf{A}^T\mathbf{KA}=\mathbf{I}}{argmax}}\:\frac{Tr(\mathbf{A}^T\mathbf{P}^{k}\mathbf{A})}{Tr(\mathbf{A}^T\mathbf{Q}^{k}\mathbf{A})},\end{equation}
%\begin{equation}\mathbf{P}^{k} = \left(\begin{array}{cccc}
%     \mathbf{P}^{k}_{11} & \mathbf{P}^{k}_{12} & ... &\mathbf{P}^{k}_{1V}\\
%     \mathbf{P}^{k}_{21} & \mathbf{P}^{k}_{22} & ... &\mathbf{P}^{k}_{2V} \\
%     ... \\
%     \mathbf{P}^{k}_{V1} & \mathbf{P}^{k}_{V2} & ... &\mathbf{P}^{k}_{VV} \end{array}\right),
%     \mathbf{Q}^{k} = \left(\begin{array}{cccc}
%     \mathbf{Q}^{k}_{11} & 0 & ... &0\\
%     0 & \mathbf{Q}^{k}_{22} & ... &0 \\
%     ... \\
%     0 & 0 & ... &\mathbf{Q}^{k}_{VV}\end{array}\right),\end{equation}
%\begin{equation}\mathbf{P}_{ij}^{k*} = \mathbf{K}_i\mathbf{L}_b\mathbf{K}_j^T, \end{equation}
%\begin{equation}\mathbf{Q}_{ii}^{k} = \mathbf{K}_i\mathbf{L}_w\mathbf{K}_i^T, \end{equation}

\begin{equation}\mathbf{P}^{k} = \mathbf{K}\mathbf{L}_b\mathbf{K}^T, \end{equation}
\begin{equation}\mathbf{Q}^{k} = \mathbf{K}\mathbf{L}_w\mathbf{K}^T, \end{equation}
where $\mathbf{L}_b$ is defined using $\mathbf{L}_{bij}^*$ or $\hat{\mathbf{L}}_{bij}$ and $\mathbf{K}$ is a block-diagonal matrix having $\mathbf{K}_v$ as its $v^{th}$ block. The solution is then given by solving the eigendecomposition problem%$\mathbf{L}_b^*$ or $\hat{\mathbf{L}}_b$ and $\mathbf{K}$ is a block-diagonal matrix having $\mathbf{K}_v$ as its $v^{th}$ block. The solution is then given by solving the eigendecomposition problem
\begin{equation}\mathbf{P}^k\mathbf{A} = \bm{\rho} \mathbf{Q}^k\mathbf{A}.\end{equation}
\subsection{Spectral Regression}
In this section, we focus on the spectral regression approach that was introduced as a way of speeding up the eigendecomposition step of LDA \cite{srda}. 
It has been shown that the solution of the generalized eigendecomposition problem (12) is equivalent to the problem $\mathbf{Jt} = \lambda \mathbf{t}$ with the same eigenpairs, for $\mathbf{t} = \mathbf{X}^T\mathbf{w}$ and $\mathbf{J}$ = $\mathbf{L}_b$:
\begin{equation}\mathbf{Jt} = \mathbf{JX}^T\mathbf{w} = \lambda \mathbf{X}^T\mathbf{w} = \lambda \mathbf{t}.\end{equation}
Exploiting this fact, the solution of (12) can be obtained by solving an eigenvalue decomposition problem $\mathbf{Jt} = \lambda \mathbf{t}$ and finding such $\mathbf{w}$ that $\mathbf{X}^T\mathbf{w} = \mathbf{t}$. In practice, such $\mathbf{w}$ may not always exist, but it can be approximated with the closest value in the least squares sense: 
\begin{equation}
%\begin{aligned}\mathbf{W} = {\underset{}{argmin}}\:(\sum_i^N(\mathbf{v}^T\mathbf{x}_i - \mathbf{t}_i)^2) = \newline \\ {\underset{}{argmin}}\:((\mathbf{X}^T\mathbf{W} - \mathbf{T})^T(\mathbf{X}^T\mathbf{W} - \mathbf{T}))\end{aligned},
\mathbf{W} = argmin\:||\mathbf{W}^T\textbf{X} - \mathbf{T}||_F^2.
\end{equation}
The solution to (36) is given by $\mathbf{W} = (\mathbf{XX}^T)^{-1}\mathbf{XT}^T$. In the cases where $\mathbf{XX}^T$ is singular, regularized solution is applied:
\begin{equation}(\mathbf{XX}^T+\alpha \mathbf{I})\mathbf{W} = \mathbf{XT}^T,\end{equation}
\begin{equation}\mathbf{W} = (\mathbf{XX}^T + \alpha \mathbf{I})^{-1}\mathbf{XT}^T,\end{equation}
where $\alpha \geq 0$ is a regularization parameter and $\mathbf{T}$ = $[\mathbf{t}_1, ..., \mathbf{t}_d]^T$.

Spectral Regression Discriminant Analysis (SRDA) was proposed as an extension to LDA based on the spectral regression \cite{srda}. It has been shown that in the case of LDA the matrix $\mathbf{J}$ (35) has $C$ eigenvectors corresponding to nonzero values, all of which correspond to the eigenvalue of 1 and have the form of  
\begin{equation}
    \mathbf{u}_i = [\underbrace{0,...,0}_{\sum_{i=1}^{p-1}N_i},\underbrace{1,...,1}_{N_p},\underbrace{0,...,0}_{\sum_{i=p+1}^CN_i}]^T,
\end{equation}
where $p$ is the class label, $N_p$ is the number of samples in class $p$ and $C$ is the number of classes. Therefore, the solution can be obtained by selecting the vector of ones as the first eigenvector and obtaining the rest by orthogonalization of the vectors of the structure as in (39). A tensor extension to SRDA has been recently proposed in \cite{hosrda}, where the eigendecomposition problem of Higher Order Discriminant Analysis is transformed into a regression problem. 

\subsection{Kernel Regression}
A kernelized version of the spectral regression was proposed in \cite{krda}. In this case, the objective is to solve the eigendecomposition problem $\mathbf{J}\mathbf{Ka} = \lambda \mathbf{Ka}$, which is equivalent to solving the eigendecomposition problem of $\mathbf{J}\mathbf{t} = \lambda \mathbf{t}$ given $\mathbf{Ka} = \mathbf{t}$:
\begin{equation}\mathbf{J}\mathbf{Ka} = \mathbf{J}\mathbf{t} = \lambda \mathbf{t} = \lambda \mathbf{Ka}.\end{equation}
Then the kernel regression is applied to obtain
\begin{equation}\mathbf{W}^* = {\underset{\mathbf{W}}{argmin}} \:||\mathbf{W}^T\bm{\Phi} - \mathbf{T}||_F^2,\end{equation}
\begin{equation}\mathbf{A} = {\underset{}{argmin}}\:||\mathbf{A}^T\bm{\Phi}^T\bm{\Phi} - \mathbf{T}||_F^2 = {\underset{}{argmin}}\:||\mathbf{A}^T\mathbf{K} - \mathbf{T}||_F^2.\end{equation}
To take into account possible singularity of $\mathbf{KK}^T$, regularized solution is used to obtain $\mathbf{A}$: 
\begin{equation}\mathbf{A} = (\mathbf{KK}^T + \alpha \mathbf{I})^{-1}\mathbf{KT}^T,\end{equation}
where $\alpha$ is the regularization parameter.
\subsection{Approximate Kernel Regression}
For large-scale datasets, kernel regression method can be substituted by an approximate kernel regression, where $\mathbf{W}$ is expressed as a linear combination of $r$ reference vectors $(r < N)$ \cite{scalecskda}. We define $\mathbf{W} = \bm{\Psi} \mathbf{A}$, where $\bm{\Psi}$ is a set of reference vectors in $\mathcal{F}$. The reference vectors in $\mathcal{F}$ correspond to $r$ prototype vectors from $\mathbb{R}^D$ that can be randomly selected training vectors from $\mathbf{X}$, random data following the same distribution as data in $\mathbf{X}$, subclass centers obtained by clustering all data, or subclass centers obtained by clustering data in each subclass separately.

Given $\mathbf{W} = \bm{\Psi}\mathbf{A}$, (42) becomes 
\begin{equation}\mathbf{A}^* = {\underset{}{argmin}}\:||\mathbf{A}^T\bm{\Psi}^T\bm{\Phi} - \mathbf{T}||_F^2 = {\underset{}{argmin}}\:||\mathbf{A}^T\mathbf{\hat{K}} - \mathbf{T}||_F^2,\end{equation} where $\mathbf{\hat{K}} = \bm{\Psi\Phi}$.
Then, \begin{equation}\mathbf{A} = (\hat{\mathbf{K}}\hat{\mathbf{K}}^T+\alpha \mathbf{I})^{-1}\hat{\mathbf{K}}\mathbf{T}^T,\end{equation}
where $\alpha$ is a regularization parameter. It should be noted that in the case $\bm{\Psi} = \bm{\Phi}$, the problem becomes equivalent to (43).

\subsection{SDA with Spectral Regression}
Subclass Discriminant Analysis has not been previously used together with Spectral Regression, but their combination is straightforward. The process of solving SDA using Spectral Regression can be defined as follows:
\begin{enumerate}
\item Create the between-class Laplacian matrix (10)
\item Solve the generalized eigendecomposition problem $\mathbf{L}_b\mathbf{t} = \lambda \mathbf{t}$ and create the matrix $\mathbf{T}$ out of the obtained vectors
\item Regress $\mathbf{T}$ to $\mathbf{W}$ as in (38) %and select the rows corresponding to the view (i.e. first $N$ rows correspond to view 1)
\item Orthogonalize $\mathbf{W}$ such that $\mathbf{W}^T\mathbf{W} = \mathbf{I}$\end{enumerate} 
Equivalently, for the kernel case, the steps 3-4 are the regression of $\mathbf{T}$ to $\mathbf{A}$ as in (43) or (45) and orthogonalization of $\mathbf{A}$ such that $\mathbf{A}^T\mathbf{KA} = \mathbf{I}$. Alternatively, the projection matrix can be $l2$-normalized instead of applying orthogonalization \cite{srda}.

The above-described process for solving the SDA optimization problem provides several advantages. Firstly, as we will show in the next section, the eigendecomposition step (35) can be substituted with a much faster process. Secondly, the eigendecomposition step (12) or (19) is avoided and substituted with the least squares regression, for which several efficient solutions exist \cite{LSQR}.

\section{Proposed approach}
In this section, the proposed methods are described. Firstly, we propose a speed-up approach for single-view SDA that relies on the structure of the Laplacian matrix $\mathbf{L}_b$ and allows to substitute the eigendecomposition step of (35) by a much faster process. Secondly, we propose a linear and kernel solutions for multi-view SDA. Thirdly, we show that the solution to multi-view SDA can be obtained by a faster process that is similar to the one described for the single-view case.

\subsection{Speeding up the eigendecomposition step}
In this section, we show how the specific block structure of the Laplacian matrix $\mathbf{L}_b$ in SDA  allows to replace the eigendecomposition step with a much faster process. 

Without loss of generality, we assume that the data in $\mathbf{X}$ is mean-centered and sorted according to the class and subclass labels, i.e., $[1, ..., N_{11}, 1,..., N_{CZ}]$, where $[1,..., N_{CZ}]$ are the subclass labels of class $C$ and subclass $Z$.

It can be observed that $\mathbf{L}_b$ has a block structure with constant values in the blocks, as described in (10), with different blocks of $\mathbf{L}_b$ corresponding to different classes. The class blocks are further divided into the subclass blocks. Since $\mathbf{L}_b$ has a block structure, its eigenvectors have the block structure as well. Moreover, bigger eigenvalues show larger differentiation and correspond to the eigenvectors discriminating class blocks, while smaller eigenvalues discriminate subblocks of class blocks, hence, representing subclasses. $\mathbf{L}_b$ has a rank of $C*Z-1$ and, therefore, it has $C*Z-1$ nonzero eigenvalues, where $Z$ is the number of subclasses in each class. 

%Due to the structure of $\mathbf{L}_b$, the block structure of the eigenvectors, that show a clear discrimination between subclasses, can be observed. 
Assuming the eigenvectors are sorted according to the eigenvalues in decreasing order, the first $C-1$ eigenvectors share similar values at indices corresponding to one class. The rest of the eigenvectors correspond to different classes, and in each of them the subclass structure of a certain class can be observed - the indices corresponding to data of the same subclass have the same nonzero value, while the indices corresponding to other classes have the value of 0. We observe that bigger eigenvalues correspond to the eigenvectors showing the subclass discrimination of classes with smaller number of samples; and the classes having the same amount of samples share the eigenvectors, i.e.,  samples at positions of both classes have nonzero values, that are the same within a subclass, while positions corresponding to other classes have the value of zero. In this case, such eigenvectors are repeated a number of times equal to the number of classes with the same amount of samples.

As an example, let us consider a problem of 2 classes, where class 1 contains 8 samples and class 2 - 9 samples. Each class contains 2 subclasses, where class 1 has 3 samples in the first subclass and 5 in the second, and class 2 has 4 samples in the first subclass and 5 samples in the second subclass. Then the three eigenvectors of the between-class Laplacian matrix of this data that correspond to nonzero eigenvalues have the structure outlined in (46), where $c$ corresponds to the class label, $z$ - to subclass label and $r_i$ - to $i^{th}$ random value. 

Moreover, $\mathbf{L}_b$ is a symmetric weightless constant sum matrix. Therefore, all of its eigenvectors are orthogonal and a vector of ones is an eigenvector with eigenvalue 0 \cite{constsum}. In addition, we can observe that for the data with a subclass structure, the eigenvectors maximizing the criterion (11) are those with the block structure as described. Following this, the orthogonalization can be performed on random vectors that follow the block structure as described above \cite{specrda}. Therefore, we can choose the vector of ones as our first eigenvector and obtain the remaining $C*Z - 1$ vectors by orthogonalizing the random vectors of the described structure following the Gram-Schmidt process \cite{qr}. As $d \leq min(D,N)$, in the case $C*Z - 1 > min(D,N)$ we can stop after $min(D,N)$ target vectors are created. The vector of ones can then be removed as being useless. The detailed process of target vectors creation is outlined in Algorithm 1.
\arraycolsep=1pt
\renewcommand{\arraystretch}{0.7}
\begin{equation}\begin{blockarray}{cccc}
\begin{block}{(ccc)c}
r_1 & r_3 &0 &  c=1, z=1\\
r_1 & r_3 &0 &  c=1, z=1\\
r_1 & r_3 &0 &  c=1, z=1\\
r_1 & r_4  &0 & c=1, z=2\\
r_1 & r_4  & 0 & c=1, z=2\\
r_1 & r_4  & 0 & c=1, z=2\\
r_1 & r_4  & 0 & c=1, z=2\\
r_1 & r_4  & 0 & c=1, z=2\\
r_2 & 0 & r_5 & c=2, z=1\\
r_2 & 0 & r_5 & c=2, z=1\\
r_2 & 0 & r_5 & c=2, z=1\\
r_2 & 0 & r_5 & c=2, z=1\\
r_2 & 0 & r_6 & c=2, z=2\\
r_2 & 0 & r_6 & c=2, z=2\\
r_2 & 0 & r_6 & c=2, z=2\\
r_2 & 0 & r_6 & c=2, z=2\\
r_2 & 0 & r_6 & c=2, z=2\\
\end{block}\end{blockarray}.\end{equation}

\begin{algorithm2e}%[H]
\SetAlgoLined
\SetKwFunction{FMain}{getSingleviewTargets}
 \SetKwProg{Fn}{Function}{:}{}
 \Fn{\FMain{$class\_labels$,$cluster\_labels$,$C$,$Z$,$N$,$D$}}{
 \KwIn{$class\_labels:$ $N\times1$ vector with class labels;
        $cluster\_labels:$ $N\times1$ vector with the cluster labels;
        $Z:$ number of clusters in each class;
        $C:$ number of classes; $N:$ number of elements; $D:$ dimensionality of data\;}
\BlankLine
$d \gets min(C*Z-1, D, N)$\;
\BlankLine
\CommentSty{\%class-level vectors}\;
\BlankLine
$T$ $\gets$ $N \times(min(d,C-1))$ matrix with random values at positions of different classes, such that values are repeated within the class in one column, but distinct between classes and columns\;

\BlankLine
$L$ $\gets$ unique numbers of elements in each class sorted in ascending order\;
\BlankLine
\CommentSty{\%cluster level vectors}\;
\BlankLine
\For{$l$ $\gets$ $\text{\upshape iterate through}$ L}{
$k$ $\gets$ classes with $l$ elements; $m$ $\gets$ length($k$)\;
\If{$size(T,2) + m*(Z-1) > d$}
{$m \gets (d-size(T,2))/(Z-1)$}
$Tclust$ $\gets$ $N$ $\times$ $m*(Z-1)$ matrix with random values at positions of all subclasses of classes in $k$, such that the values are shared within the subclass in one column, but distinct between subclasses and columns. Values at positions of other classes are 0s\; 
$T$ $\gets$ append $Tclust$ as columns on the right\;
\If{$size(T,2) == d$}{\textbf{break}}
}
\BlankLine

$T$ $\gets$ append $N$$\times$$1$  vector of ones as a column on the left\;
Orthogonalize $T$; remove first column of $T$;

\KwRet{$T^T$}
}
\caption{Target vectors calculation, single-view case}
\end{algorithm2e}

\subsection{Multi-view Subclass Discriminant Analysis}
In this section, we propose a novel method for multi-view subspace learning - Multi-view Subclass Discriminant Analysis along with the kernelized version. Unlike previously described SMvDA and MvMDA methods that similarly to LDA assume that data within each view follows a unimodal Gaussian distribution, we propose a method that would take into account potential multimodalities present in the data of each view and hence lead to a more robust solution. This is done by modelling data of each view with multiple subclasses. Therefore, the idea behind multi-view Subclass Discriminant Analysis is the maximization of the distance between the subclass means of different classes, while minimizing the distances between the samples of the same subclass. The total scatter matrix for the mean-centered data is defined as  
\begin{equation}\mathbf{S}_t = \sum_{i=1}^V\sum_{k=1}^{N}\mathbf{y}_k^i{\mathbf{y}_k^i}^T = \mathbf{Y}\mathbf{Y}^T = \mathbf{W}^T\mathbf{XX}^T\mathbf{W}, \end{equation}
where $\mathbf{y}_k^i$ is the $k^{th}$ sample of view $i$ in the latent space. The between-class scatter matrix is defined as 
\begin{equation}\begin{aligned}
\mathbf{S}_b = {} &\sum_{i=1}^{V}\sum_{j=1}^{V}\sum_{p=1}^{C}\sum_{\underset{q\ne p}{q=1}}^{C}\sum_{l=1}^{d_p}\sum_{h=1}^{d_q}p_{pl}^ip_{qh}^j(\bm{\mu}_{pl}^i - \bm{\mu}_{qh}^j)(\bm{\mu}_{pl}^i - \bm{\mu}_{qh}^j)^T \\
& = \sum_{i=1}^{V}\sum_{j=1}^{V}\mathbf{W}_i^T\mathbf{X}_i\mathbf{L}_{bij}^{mv}\mathbf{X}_{j}^T\mathbf{W}_j = \mathbf{W}^T\mathbf{XL}_b^{mv}\mathbf{X}^T\mathbf{W},\end{aligned}\end{equation}
\begin{equation}\mathbf{X} = \left(\begin{array}{cccc}
    \mathbf{X}_1 & 0 & ... & 0 \\
    0 & \mathbf{X}_2 & ... & 0 \\
    0 & 0 & ... & \mathbf{X}_v \end{array}\right),\end{equation}
    \begin{equation}\mathbf{W} = \left(\begin{array}{c}
    \mathbf{W}_1\\
    \mathbf{W}_2 \\
    ... \\
    \mathbf{W}_v \end{array}\right),\end{equation}
    
\renewcommand{\arraystretch}{1.2}    
\begin{equation}\mathbf{L}_b^{mv} = \left(\begin{array}{cccc}
     \mathbf{L}_{b11}^{mv}  & \mathbf{L}_{b21}^{mv} & ... & \mathbf{L}_{bV1}^{mv}\\
     \mathbf{L}_{b12}^{mv}  & \mathbf{L}_{b22}^{mv} & ... & \mathbf{L}_{bV2}^{mv}  \\
     ... & ... & ... & ...  \\
     \mathbf{L}_{b1V}^{mv}  & \mathbf{L}_{b2V}^{mv} & ... & \mathbf{L}_{bVV}^{mv}  \end{array}\right),\end{equation}
\small
     \begin{equation}\mathbf{L}_{bij}^{mv} = \begin{cases}
			2\sum_{p=1}^{C}\sum_{\underset{q\ne p}{q=1}}^{C}\sum_{l=1}^{d_p}\sum_{h=1}^{d_q}\frac{VN_{qh}^j}{N_{pl}^iN^2}\mathbf{e}_{pl}^i {\mathbf{e}_{pl}^i}^T - \frac{1}{N^2}\mathbf{e}_{pl}^i{\mathbf{e}_{qh}^j}^T,&\text{if}$\ $i=j\\
            -2\sum_{p=1}^{C}\sum_{\underset{q\ne p}{q=1}}^{C}\sum_{l=1}^{d_p}\sum_{h=1}^{d_q}\frac{1}{N^2}\mathbf{e}_{pl}^i{\mathbf{e}_{qh}^j}^T, & \text{otherwise}
		 \end{cases},\end{equation}
		 \normalsize
where $i$ and $j$ are view labels, $p$ and $q$ are class labels, $l$ and $h$ are subclass labels, $p_{pl}^i = \frac{N_{pl}^i}{N}$ is the prior of the subclass $l$ of class $p$ in the view $i$, $\bm{\mu}_{pl}^i$ is the mean of the subclass $l$ of class $p$ in view $i$, $\mathbf{e}_{pl}^i$ is the vector of length $N$ with ones at positions corresponding to subclass $l$ of class $p$ in view $i$ and zeros elsewhere.
		 
	The solution is then obtained by optimizing the Fisher-Rao's criterion:
\begin{equation}
%    J = {\underset{W_i, i=1,...,v}{argmax}}\frac{Tr(\sum_i^V\sum_j^VW_i^TX_iL_bX_j^TW_j)}{Tr(\sum_i^VW_i^TX_iX_i^TW_i)}
    \mathcal{J}(\mathbf{W}) = {\underset{\mathbf{W}_i^T\mathbf{W}_i = \mathbf{I}, i=1,...,V}{argmax}}\:\frac{Tr(\mathbf{W}^T\mathbf{XL}_b\mathbf{X}^T\mathbf{W})}{Tr(\mathbf{W}^T\mathbf{XX}^T\mathbf{W})} ,
\end{equation}
where $\mathbf{X}$ and $\mathbf{W}$ are defined as in (49) and (50), respectively, and $\mathbf{K}$ is centered. Equivalently, solution to the kernel version of the method is obtained by optimizing \begin{equation}
%    J = {\underset{W_i, i=1,...,v}{argmax}}\frac{Tr(\sum_i^V\sum_j^VW_i^TX_iL_bX_j^TW_j)}{Tr(\sum_i^VW_i^TX_iX_i^TW_i)}
    \mathcal{J}(\mathbf{A}) = {\underset{\mathbf{A}^T\mathbf{KA} = \mathbf{I}}{argmax}}\frac{Tr(\mathbf{A}^T\mathbf{KL}_b\mathbf{K}^T\mathbf{A})}{Tr(\mathbf{A}^T\mathbf{KK}^T\mathbf{A})},
\end{equation}
where $\mathbf{K}$ is a block-diagonal matrix having $\mathbf{K}_v$ as its $v^{th}$ block.

%\begin{equation}K = \left(\begin{array}{cccc}
%    K_1 & 0 & ... & 0 \\
%    0 & K_2 & ... & 0 \\
%    0 & 0 & ... & K_v \end{array}\right)\end{equation}

The solution to (53) is obtained by solving the eigendecomposition problem $\mathbf{L}_b\mathbf{X}^T\mathbf{v} = \lambda \mathbf{X}^T\mathbf{v}$. Similarly, the solution to (54) is given by  $\mathbf{L}_b\mathbf{Ka} = \lambda \mathbf{Ka}$. Both of these problems can be solved by the process equivalent to the one described in 3.1. 
\begin{algorithm2e}%[H]
\SetAlgoLined
\SetKwFunction{FMain}{getMultiviewTargets}
 \SetKwProg{Fn}{Function}{:}{}
 \Fn{\FMain{$class\_labels$,$cluster\_labels$,$V$,$C$,$Z$,$N$,$D$}}{
 \KwIn{$class\_labels:$ $V*N\times1$ vector with class labels;
        $cluster\_labels:$ $V*N\times1$ vector with the cluster labels;
        $V:$ number of views;
        $Z:$ number of clusters in each class;
        $C:$ number of classes; $N:$ number of elements; $D:$ vector of dimensionalities of data in each view\;}
\BlankLine
$d \gets min(V*C*Z-1, min(D), N)$\;
\BlankLine
\CommentSty{\%class-level vectors}\;
\BlankLine
$T$ $\gets$ $V*N\times(min(d,C-1))$ matrix with random values at positions of different classes, such that values are repeated within the class in one column, but distinct between views, classes, and columns\;
%\For{$c\gets1$ \KwTo C }{
%Put random values to Y for all rows of class c, such that values are the same in each column, but different in each row\;
%}
%Add $N$\times$1$ vector of ones to Y\;
%Orthogonalize Y via Gram-Schmidt process; remove first column of Y\;
%Append Y as new columns to T on the right\; %T = [T Y];
\BlankLine
$L$ $\gets$ unique numbers of elements in each class sorted in ascending order\;
\BlankLine
\CommentSty{\%cluster level vectors}\;
\BlankLine
\For{$l$ $\gets$ $\text{\upshape iterate through}$ L}{
$k$ $\gets$ classes with $l$ elements; $m$ $\gets$ length($k$)\;
\If{$size(T,2) + m*(V*Z-1) > d$}
{$m\gets(d-size(T,2))/(V*Z-1)$}
$Tclust$ $\gets$ $V*N$ $\times$ $m*(V*Z-1)$ matrix with random values at positions of all subclasses of classes in $k$, such that the values are shared within the subclass in one column, but distinct between subclasses, views, and columns. Values at positions of other classes are 0s\; 
$T$ $\gets$ append $Tclust$ as columns on the right\;
\If{$size(T,2) == d$}{\textbf{break}}
}
\BlankLine

$T$ $\gets$ append $N\times1$ vector of ones as a column on the left \;
Orthogonalize $T$; remove first column of $T$;

\KwRet{$T^T$}
}
\caption{Target vectors calculation, multi-view case}
\end{algorithm2e}

\subsection{Speeding up the eigendecomposition step: multi-view case}
In this section, we describe a speed-up approach for the Multi-view Subclass Discriminant Analysis, based on the specific structure of the Laplacian matrix $\mathbf{L}_b^{mv}$. The process of speeding up the eigendecomposition step for the multi-view case is similar to the single-view one. The Laplacian matrix $\mathbf{L}_b^{mv}$ is the constant sum symmetric block matrix, thus having orthogonal eigenvectors, one of which is the vector of ones corresponding to eigenvalue of 0. The matrix has a block structure, where different blocks correspond to different views, and inside of each diagonal view block we can observe the block structure that is the same as in the single-view case. Due to this block structure the eigenvectors of $\mathbf{L}_b^{mv}$ have the block structure as well. Assuming that the number of clusters is the same in all views, the rank of the $\mathbf{L}_b^{mv}$ is $C*Z*V-1$, and that is the maximum number of nonzero eigenvalues.

Let us consider the data of 2 views and 2 classes. Let  1 contain 2 subclasses, with 2 samples in the first subclass and 2 samples in the second subclass in both views. Let class 2 contain 2 subclasses, with 3 samples in the first subclass and 4 samples in the second subclass in the first view, and 4 samples in the first subclass and 3 samples in the second subclass in the second view. Then the eigenvectors of the between-class Laplacian matrix of the proposed multi-view SDA will have the structure outlined in (55), where $c$ corresponds to the class label, $z$ corresponds to the subclass label, $v$ corresponds to the view label and $r_i$ corresponds to $i^{th}$ random value.
\renewcommand{\arraystretch}{0.7}

\begin{equation} 
\arraycolsep=0.1pt \begin{blockarray}{cccccccccc}
& & & & & & & c&z& v\\
\begin{block}{(ccccccc)ccc}
% & & & & & & & c&z& v\\
r_1 & r_5 & r_9& r_{13} & 0& 0& 0 & 1&1& 1\\
r_1 & r_5 & r_9&r_{13} & 0& 0&  0 & 1&1& 1\\
r_1 & r_6& r_{10}&r_{14}  & 0& 0 & 0 & 1&2& 1\\
r_1 & r_6 & r_{10}&r_{14}  & 0& 0 & 0 & 1&2& 1\\
r_2 & 0& 0 &0 & r_{17}& r_{21} & r_{25} & 2&1& 1\\
r_2 & 0 & 0&0 & r_{17}& r_{21} & r_{25} & 2&1& 1\\
r_2 & 0 & 0&0 & r_{17} & r_{21}& r_{25} & 2&1& 1\\
r_2 & 0 & 0&0 & r_{18} & r_{22}& r_{26} & 2&2& 1\\
r_2 & 0 & 0&0 & r_{18} & r_{22}& r_{26} & 2&2& 1\\
r_2 & 0 & 0&0 & r_{18} & r_{22}& r_{26} & 2&2& 1\\
r_2 & 0 & 0&0 & r_{18} & r_{22}& r_{26} & 2&2& 1\\
r_3 & r_7& r_{11} &r_{15} & 0& 0&  0 & 1&1& 2\\
r_3 & r_7& r_{11} &r_{15}  & 0& 0& 0 & 1&1& 2\\
r_3 & r_8& r_{12} &r_{16}  & 0& 0& 0 & 1&2& 2\\
r_3 & r_8& r_{12} &r_{16}  & 0& 0 & 0 & 1&2& 2\\
r_4 & 0 & 0&0 & r_{19} & r_{23}& r_{27} & 2&1& 2\\
r_4 & 0 & 0&0 & r_{19} & r_{23}& r_{27} & 2&1& 2\\
r_4 & 0 & 0&0 & r_{19} & r_{23}& r_{27} & 2&1& 2\\
r_4 & 0 & 0&0 & r_{19} & r_{23}& r_{27} & 2&1& 2\\
r_4 & 0& 0&0 & r_{20} & r_{24}& r_{28} & 2&2& 2\\
r_4 & 0& 0 &0 & r_{20} & r_{24}& r_{28} & 2&2& 2\\
r_4 & 0 & 0&0 & r_{20} & r_{24}& r_{28} & 2&2& 2\\
\end{block}\end{blockarray}.\end{equation}

It can be observed that the first $C-1$ eigenvectors have the class block structure similar to the one in the single-view case, and the blocks are repeated across the positions corresponding to the different views. In the same way as in the previously described single-view case, the rest of the eigenvectors correspond to different classes and each of them exposes the subclass structure of specific class - the values corresponding to the same subclass are the same within each view in the eigenvector and the values corresponding to other classes are 0 in all the views. We observe that the classes with the same amount of samples share the eigenvectors in a similar way to the single-view case, and these eigenvectors are repeated for the number of times equal to the number of classes sharing the number of elements. The eigenvectors showing subclass discrimination of smaller classes correspond to bigger eigenvalues.

Following the procedure described for single-view case, the eigenvectors can be obtained by forming the random vectors of the structure described, and orthogonalizing starting from the vector of ones following the Gram-Schmidt process \cite{qr}. The vector of ones can then be removed. The detailed procedure is described in Algorithm 2.

\subsection{Computational Complexity Analysis}
In this section, we discuss the complexity analysis of the original Subclass Discriminant Analysis and the proposed speed-up approach. The complexities are described using $flam$ - a compound operation denoting one addition and one multiplication \cite{ruhe2000matrix}. Complexity of SDA can be defined as follows:
\begin{itemize}
    \item Calculation of total scatter matrix $\mathbf{S}_t$ is $\frac{ND^2}{2} + DN$
    \item Calculation of between-class scatter matrix $\mathbf{S}_b$ is $\frac{Z^2C(C-1)}{2}(\frac{D(D+1)}{2}+ 1 + D) + 2DN$, where $Z$ is the number of subclasses and $C$ is a number of classes
    \item Solving the eigendecomposition problem in (8) is $\frac{9}{2}D^3 +  2D^3 + D^3 = \frac{15}{2}D^3$ \cite{srda, ruhe2000matrix}
\end{itemize}
It should be noted here that we do not perform the calculation of the stability criterion defined in \cite{sda} as the number of subclasses is defined explicitly for fair comparison with other methods - fixing the same cluster labels for data points in all the methods ensures that clustering accuracy plays no effect on overall accuracy and fine-tuning the number of clusters does not affect the speed. Besides, implementation following the $SDA_{stab}$ algorithm requires multiple iterations with different subclass numbers therefore making the algorithm more computationally intensive.

The computational complexity of fastSDA depends on three steps: mean-centering of data, creation of target vectors, and regression step. Let us first consider the regression problem in (38). First, we can notice that for a positive definite matrix $(\mathbf{XX}^T + \alpha \mathbf{I})$, inversion can be performed efficiently via Cholesky decomposition, i.e., $(\mathbf{XX}^T + \alpha \mathbf{I})^{-1} = (\mathbf{R}^{-1})(\mathbf{R}^{-1})^T$, where $\mathbf{R}$ is an upper triangular matrix, s.t., $(\mathbf{XX}^T + \alpha \mathbf{I}) = \mathbf{R}^T\mathbf{R}$. Further, when $N < D$, the problem in (38) can be transformed into $\mathbf{X}(\mathbf{X}^T\mathbf{X})^{-1}\mathbf{T}^T$.
Thus, the complexity of fastSDA can be calculated as follows:

\begin{itemize}
\item Complexity of target vector creation is essentially equal to complexity of Gram-Schmidt process = $Nd^2 - \frac{1}{3}d^3$, where $d = ZC-1$ is the dimensionality of the projection space (number of target vectors), $d < D$ \cite{srda, ruhe2000matrix}.
\item Complexity of mean-centering data is $DN$
\item If $N \leq D$, complexity of the regression step is $\frac{DN^2}{2} + \frac{N^3}{6} + dN^2+DNd$, where $\frac{DN^2}{2}$ is the complexity of $\mathbf{X}^T\mathbf{X}$, $\frac{N^3}{6}$ is the complexity of Cholesky decomposition \cite{ruhe2000matrix, srda}, and $d^2N+DNd$ is the complexity of the remaining regression steps (i.e., calculation of inverse and multiplication). 
\item If $N > D$, complexity is $\frac{ND^2}{2}+ \frac{D^3}{6} + dD^2 + DNd$, where $\frac{ND^2}{2}$ is the complexity of $\mathbf{XX}^T$, $\frac{D^3}{6}$ is the complexity of Cholesky decomposition \cite{ruhe2000matrix, srda}, and $d^2D+DNd$ is the complexity of the remaining regression steps (i.e., calculation of inverse and multiplication).
\item Normalization of $\mathbf{W}$ has complexity of $\frac{d^2D}{2} + dD + d$
\end{itemize}

\begin{figure}[h!]
    \centering
        \includegraphics[width=0.95\textwidth]{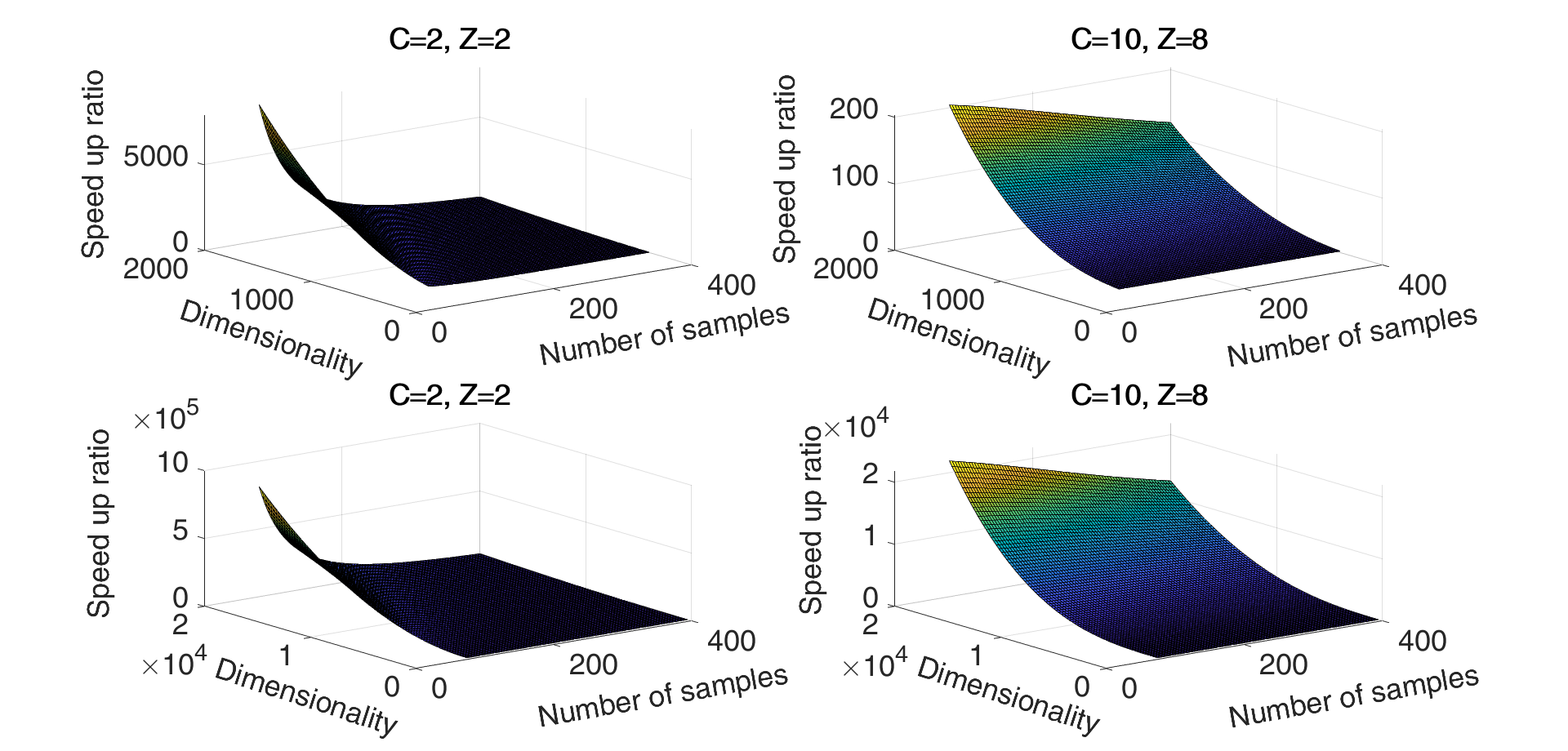}
        \caption{Dependancy of speed up ratio based on $N,D,C,Z$ for $N < D$.}
        \label{dntime}
\end{figure}

In the case $N<D$, the most computationally intensive term of fastSDA depends on relation of $d$ to $N$. It is equal to $\frac{DN^2}{2}$ for $N>2d$ and $DNd$ for $N<2d$. Both of these terms are smaller than $D^3$ as both $N$ and $d$ are smaller than $D$. In turn, the computational complexity of SDA is at least $\frac{15}{2}D^3$. Therefore, fastSDA will always outperform SDA in this scenario. The dependency of speed-up on dimensionality and the number of samples can be seen from Fig.~\ref{dntime}. Here we show the ratio of SDA training time to fastSDA training time for four different cases of $D, N, C, Z$, where we consider cases of large and small number of classes/subclasses for larger/smaller $D$. As can be observed, the speed up ratio increases with the increase of $D$. 

\begin{figure}[h!]
    \centering
        \includegraphics[width=0.9\textwidth]{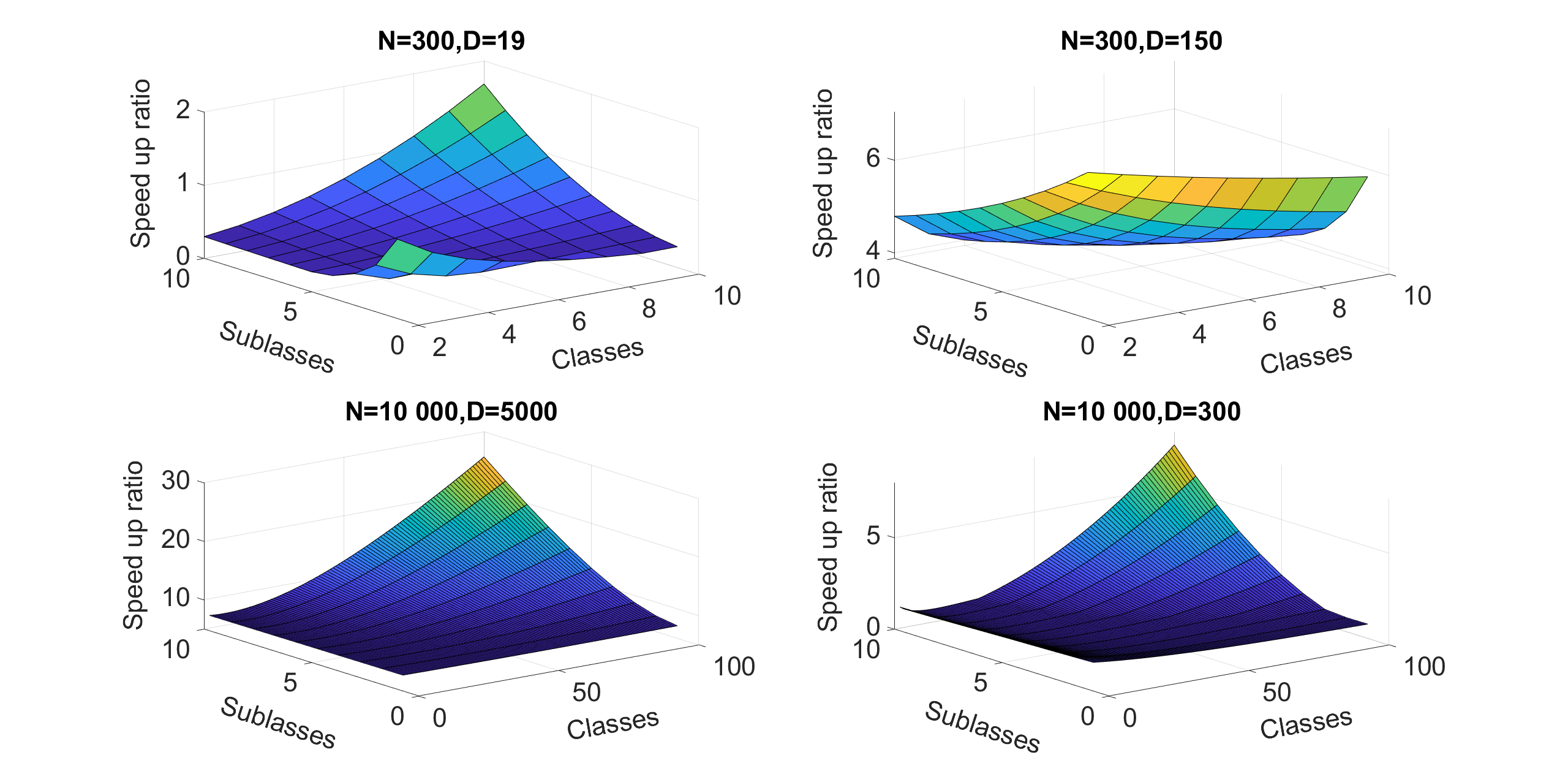}
        \caption{Dependancy of speed up ratio based on $N,D,C,Z$ for $N > D$.}
        \label{cztime}
\end{figure}

In the case $N>D$, the highest term of SDA is at least $\frac{ND^2}{2}$ if $N>15D$ and $\frac{15}{2}D^3$ otherwise. Besides, the complexity term of calculation of between-class scatter becomes significant for higher $C$ and $Z$. For fastSDA the largest term is $\frac{ND^2}{2}$ if $d<\frac{D}{2}$ and $DNd$ otherwise. Thus, the speed-up ratio depends on the five parameters of $D,N,C,Z,d$. The dependency of training time on $C$ and $Z$ is shown on Fig.~\ref{cztime} where four cases are considered: two for a small dataset and two for a large dataset, where in each case we show the subcases where $N > 15D$ and $N<15D$ and the ratio of training time of SDA to that of fastSDA is shown. It can be observed that for big enough $D$ fastSDA always results in a speed up that becomes more significant with higher $D$, $C$, and $Z$. For the case where $D$ is small, $D << N$, and both $C$ and $Z$ are small, the speed-up might not always be achieved or be close to 1, as can be observed from the top-left subplot. However, we argue that scenario where $D << N$ is not a common case where dimensionality reduction is applied in the first place. Experimentally we also show that our methods results in superior speed for both large-scale datasets with many classes (e.g., SoF) and small low-dimensional datasets with lower number of classes (e.g., Ionosphere, Monks2).

In the kernel case, the regression step (43) is equivalent to $(\mathbf{K} + \alpha\mathbf{I})^{-1}\mathbf{T}^T$ for normalized $\mathbf{A}$. The complexity of kernel fastSDA can then be calculated as follows:
\begin{itemize}
    \item Complexity of kernel matrix calculation is $DN^2$
    \item Complexity of mean-centering of $\mathbf{K}$ is $\frac{N^2}{2} + \frac{N^3}{2}$
    \item Complexity of target vectors creation is $Nd^2 - \frac{1}{3}d^3$
    \item Complexity of regression step is $\frac{N^3}{6} + dN^2$
    \item Complexity of normalization of the projection matrix is $dN^2 + d^2N + dN + d$
\end{itemize}
For kernel SDA, the step of kernel matrix creation is added as well. Besides, as kernel SDA formulation is based on graph embedding framework, the complexity of KSDA becomes at least equal to the complexity of eigendecomposition (19) which is equal to $\frac{15}{2}N^3$, while the largest term of fastSDA complexity becomes $\frac{2}{3}N^3$ (here we exclude the kernel matrix creation term as it is equal for both methods). The complexity of multi-view fastSDA can be computed similarly, following $N = \sum_{i=1}^VN_i$ and $D = \sum_{i=1}^VD_i$ ($D = \sum_{i=1}^VN_i$ in the kernel case), where $V$ is the number of views.

We can conclude that fastSDA outperforms SDA in terms of computational complexity and the speed-up increases with the dimensionality of data. Besides, another speed-up factor comes from the computation of the between-class scatter matrix that becomes much more computationally intensive with larger number of classes and/or subclasses in SDA. 

\section{Experimental results}
In this section, the experimental results are presented. The results are compared with other subspace learning techniques, namely SDA, CDA, SMFA, and SRDA, as well as the kernel SDA, CDA, and SMFA. After feature extraction, classification is performed with $k$-Nearest Neighbors classifier with $k$ = 5. In addition, we verify some of the assumptions regarding the proposed approach by performing eigendecomposition of $L_b$, regressing the obtained eigenvectors following (38) and projecting the data onto the obtained vectors that correspond to larger criterion values (11). 

For the kernel version of the methods, we exploit the RBF kernel function: 
\begin{equation}
    \mathbf{K}(x_i,x_j) = exp(-\frac{||\mathbf{x}_i - \mathbf{x}_j||_2^2}{2\sigma^2}),
\end{equation}
where we set the Gaussian scale $\sigma$ to the mean Euclidean distance between the training vectors. 

In our experiments, we assume that the subclass label of each data point in each class is known and is determined by applying the k-means clustering in $\mathbb{R}^D$. The performance is tested for the different numbers of clusters $Z = \{1,2,3,4,5,6\}$, and the same number of clusters is used for each class. We perform clustering in the original space and use the same cluster labels in the kernel methods. The same subclass labels are used for all subclass-based methods to guarantee that the differences in performance observed between the methods are not related to the specific clustering solutions of K-Means, but on the optimization problem each method adopts for determining the corresponding subspace. In the multi-view case, data in each view is clustered separately. The dimensionality $d$ of the projection space is defined by the rank of the $\mathbf{L}_b$ or $\mathbf{L}_b^{mv}$ matrix and is equal to $C*Z - 1$ and $V*C*Z - 1$, respectively, where $V$ is the number of views, $C$ is the number of classes, and $Z$ is the number of clusters. 

For each experiment, 5-fold stratified cross-validation was used, with 60\% of data of each class belonging to training set, 20\% to validation set, and 20\% to test set, where the validation set is used for hyperparameter tuning, and results are reported by training on the training set and testing on the test set. All experiments were performed on a computer with 4-core Intel i7-4800Q CPU and 32 GB of RAM.

For single-view approaches, prior to using any method, we applied PCA, preserving the eigenvectors corresponding to 98\% of the total energy and the data was standardized. The hyperparameters of all methods, if any, were tuned with the grid search. For SMFA, $k_{Int}$ and $k_{Pen}$ were selected from the range of $[2,14]$ with step 3 and $[20,100]$ with step 20, respectively. 

For calculating the distance matrix in SMFA/KSMFA, Gaussian similarity (56) with $\sigma$ equal to the mean Euclidean distance between the training vectors was used. The regularization parameter for the kernel regression was chosen from the set $[1e^{-3}, 1e^{-2}, ..., 1e^3]$. For the regularization of the other single-view kernel methods and multi-view methods the same parameter range was used. Cholesky decomposition was used for efficient matrix inversion.

In the multi-view kernel case, the solutions for the datasets containing more than 2500 samples were obtained with approximate kernel regression with the kernel matrix formed with 1500 random vectors from the training data. In the single-view kernel case, the approximate kernel regression \cite{scalecskda} with the prototype vectors formed by clustering all data with cardinality of 1000 was used on a large-scale SoF dataset for the proposed approach. On this dataset, the Nyström-based approximate kernel \cite{aksl} was used for KSDA, KCDA, and KSMFA methods with cardinality 1000. 

\subsection{Single-view datasets}

\begin{figure}[h!]
    \centering
        \includegraphics[width=\textwidth]{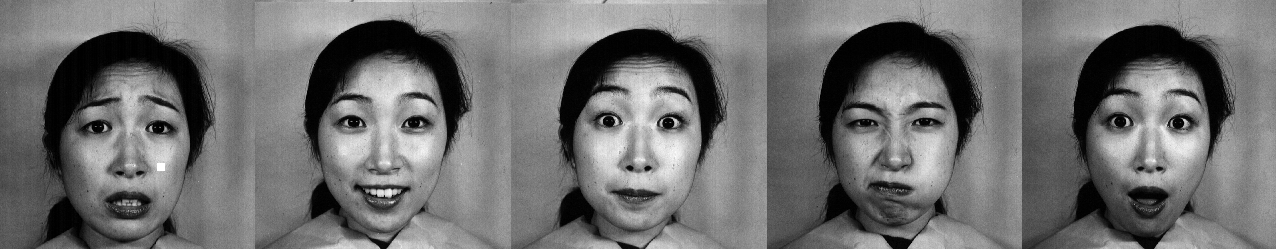}
        \caption{An example of images from Jaffe dataset.}
        \label{jaffe}
\end{figure}

We conducted experiments on 4 facial image datasets, one large-scale facial image dataset, and 4 other datasets of various data types.
The Jaffe \cite{jaffe} dataset contains facial images of Japanese females with 7 different facial expressions: anger, happiness, fear, disgust, sadness, surprise, and neutral. The dataset consists of 213 images and several examples can be seen from Fig.~\ref{jaffe}.

BU \cite{bu} dataset contains 700 images of individuals with the same 7 facial expressions. 

\begin{figure}[h!]
    \centering
        \includegraphics[width=\textwidth]{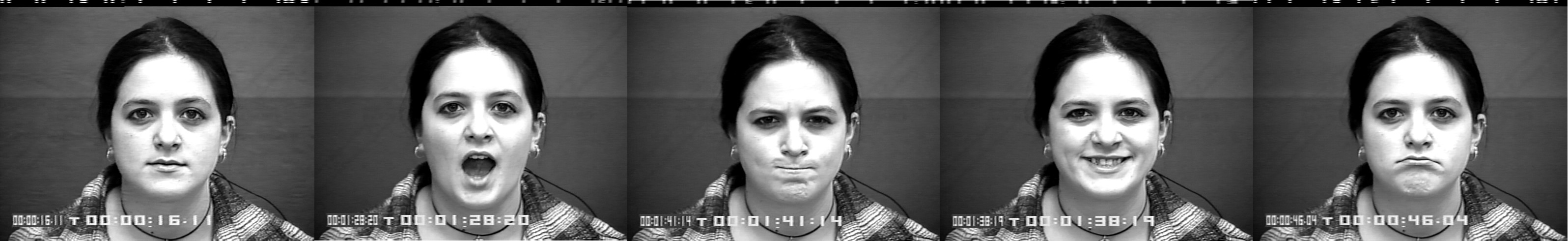}
        \caption{Examples of images from Cohn-Kanade dataset.}
        \label{kanade}
\end{figure}

The Cohn-Kanade \cite{kanade} dataset contains 245 images of different people with different facial expressions of the same 7 classes as BU and Jaffe datasets. Example images can be seen in Fig.~\ref{kanade}.

\begin{figure}[h!]
    \centering
        \includegraphics[width=\textwidth]{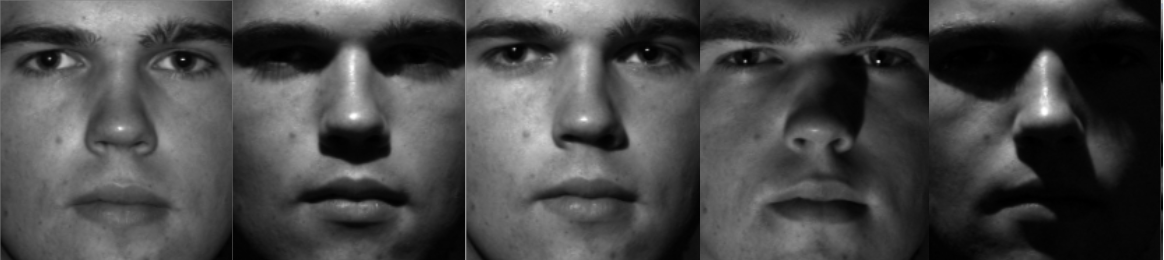}
        \caption{Examples of images from Yale dataset.}
        \label{yale}
\end{figure}

The Extended Yale-B dataset \cite{yale} contains 2432 grayscale facial images of 38 people and, therefore, defines a face recognition problem with 38 classes. Each class is represented by 64 images of the same person under different illumination conditions, positions, and view angles. Example images can be seen in Fig.~\ref{yale}.

\begin{figure}[h!]
    \centering
        \includegraphics[width=\textwidth]{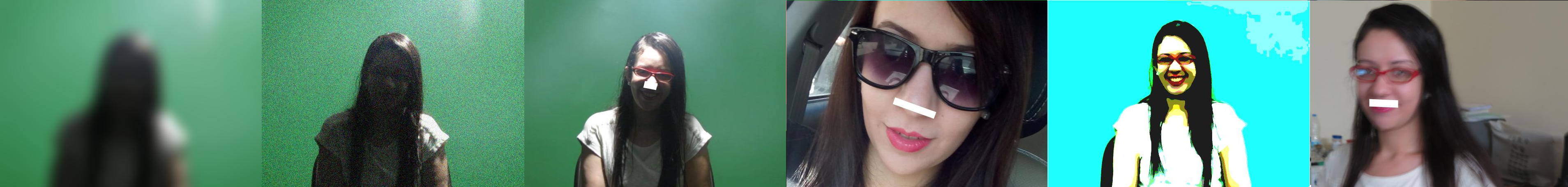}
        \caption{Examples of images from SoF dataset.}
        \label{sof}
\end{figure}

The large-scale SoF dataset \cite{SoF} consists of 42,592 images of 112 persons (66 male and 46 female) collected under different illumination conditions and containing images with occlusions (e.g. glasses). Example images can be seen in Fig.~\ref{sof}. All the facial image datasets mentioned above (i.e., Jaffe, Cohn-Kanade, BU, Yale-B, SoF) were reshaped to images of $30\times40$ pixels and flattened to obtain $1200\times1$ vectors. 

The Ionosphere dataset \cite{ionosphere} contains radar data represented as 351 34-dimensional vectors, along with the information on whether they contain evidence of some type of structure in the ionosphere or not, hence posing a binary classification problem. The Semeion dataset \cite{semeion} contains 1593 instances of handwritten digits produced by 80 persons, each of whom had written each digit twice, – in a normal way and in a fast way. The digits are represented by 16x16 binarized images flattened to $256\times1$ vectors. 

The MONKS2 dataset \cite{monks} is derived from a domain, where each instance is represented by 6 discrete features corresponding to one of the two classes. The artificially generated data describes certain physical properties of robots, and the task is to predict the type of the robot based on these characteristics. The PIMA Indians Diabetes dataset \cite{PIMA} contains information on various medical attributes of patients, including the number of pregnancies the patient has had, their BMI, insulin level, age, along with the information on whether the patient has diabetes. The dataset contains 768 instances.

\subsection{Multi-view datasets}
For the evaluation of the multi-view methods seven datasets were used: Handwritten digits \cite{handwritten}, Caltech-101 \cite{caltech, gitdata}, NUS-WIDE \cite{nuswide, gitdata}, Human Action Recognition Using Smartphones \cite{har}, Robots Execution Failures \cite{robots}, Healthy Old People Action Recognition \cite{oldpeople}, Million Song Dataset with Images (MSDI) \cite{msdi}.
The Handwritten digits dataset (HWD) \cite{handwritten} contains 2000 instances of handwritten digits of 10 classes. The images are represented by 6 views: Fourier coefficients $(1\times128)$, profile correlations $(1\times76)$, Karhunen-Love coefficients $(1\times64)$, pixel averages $(1\times240)$, Zernike moments $(1\times47)$, and morphological features $(1\times6)$.

\begin{figure}[h!]
    \centering
        \includegraphics[width=\textwidth]{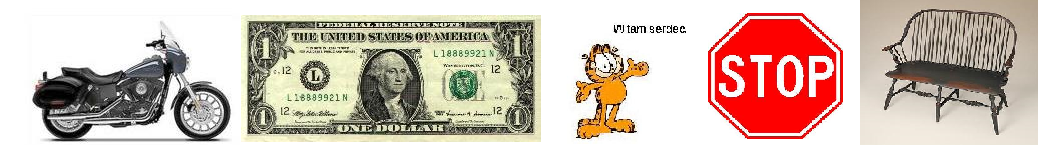}
        \caption{Examples of images from Caltech-101 dataset.}
        \label{caltech}
\end{figure}

The Caltech-101 dataset \cite{caltech} is an image classification dataset represented by 6 views: Gabor features $(1\times48)$, wavelet moments $(1\times40)$, CENTRIST features $(1\times254)$, Histogram of oriented gradients features $(1\times1984)$, GIST features $(1\times512)$, local binary pattern features $(1\times928)$. Due to imbalanced data between classes, the dataset is divided into two subsets of 7 and 20 classes, resulting in 1474 and 2386 instances, respectively. Examples of images can be seen in Fig.~\ref{caltech}.

\begin{figure}[h!]
    \centering
        \includegraphics[width=\textwidth]{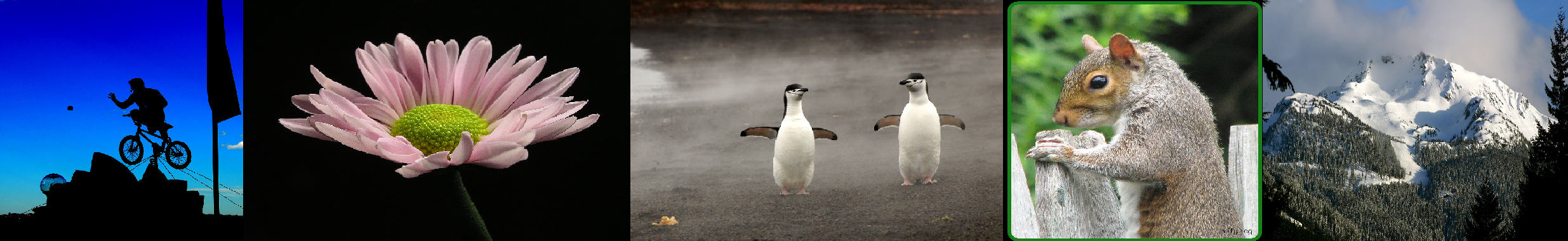}
        \caption{Examples of images from NUS-WIDE dataset.}
        \label{nuswide}
\end{figure}

NUS-WIDE dataset \cite{nuswide} is a large-scale image classification dataset of 31 classes described from 5 views: color histogram $(1\times65)$, color moments $(1\times226)$, color correlation $(1\times145)$, edge distribution $(1\times74)$, wavelet texture $(1\times129)$. Due to the large amount of samples, a subset of 11288 instances is selected for the experiments. Examples of images can be seen in Fig.~\ref{nuswide}.

The Human Action Recognition Using Smartphones dataset (HARS) \cite{har} contains  3-axial angular velocity and linear acceleration data taken from the accelerometer and gyroscope data of a smart phone attached to a person's waist while the person is performing one of the 6 activities. Actions are described from 9 views:  angular velocity of each of 3 axes, total acceleration of each of 3 axes, and body acceleration of each axis. Each view has the dimensionality of 128. Data was gathered from a group of 30 volunteers, resulting in 7352 instances. The cross-validation splits in our experiments were done such that the subjects performing the experiments are not repeated between training, validation, and test splits. 
 
Healthy Old People Action Recognition dataset (HOPAR) \cite{oldpeople} contains 2 datasets, each containing the information from a wireless sensor worn by a person, while performing one of the 4 activities: sitting on a bed, sitting on a chair, lying, ambulating. The data is organized into 4 views, where views 1-3 represent the acceleration from each of the 3 axes and view 4 contains information about the received signal strength indication, frequency, and phase of the signal, obtained from the sensor. The first dataset consists of data obtained from 60 subjects, out of which 25\% of each class instances were selected, resulting in 10495 instances. The second dataset contains information obtained from 27 subjects, resulting in 9057 instances.

The Robot Execution Failures dataset \cite{robots} consists of 5 subsets, each describing a different problem. For our experiments, subsets 1 and 4 were combined, resulting in a dataset of failures in approach to grasp or ungrasp position. The data is represented by 4 classes: normal, collision, frontal collision and obstruction, and described from 6 views: force on each of the 3 axes and torque on each of the 3 axes. Each view has 15 dimensions. The dataset consists of 205 instances.

The Million Song Dataset with Images (MSDI) \cite{msdi} poses a music genre classification task for 15 different genres. Each instance represents a song, that is described from two views: audio spectrograms from audio signal and CNN features of the corresponding album cover. Both views have the dimensionality of 200. We perform evaluation on the subset of 7468 instances, chosen randomly from the dataset and preserving the initial class proportions. 

\subsection{Results}
\renewcommand{\arraystretch}{1.3}

Tables 1 and 2 show the results for the single-view linear methods, where Table 1 depicts the accuracy and the number of subclasses resulting in the best accuracies, and Table 2 shows the training time in seconds. Tables 3 and 4 show similar information for the kernel methods. We performed experiments on 9 datasets using the proposed approach, which is compared to the conventional eigendecomposition-based approaches of SDA, CDA, SMFA and with SRDA in the linear case; and KSDA, KCDA and KSMFA for the kernel case.

Tables 5 and 6 show the results for the multi-view case in the linear formulations, while tables 7 and 8 show the same information for kernel formulations. The results are presented similarly to Tables 1 and 2. The following methods are compared: single-view SDA, where features from different views are concatenated, MvMDA, and SMvDA. For the single-view SDA we use the proposed fast approach. We report the accuracy, time taken for training, and the number of subclasses that resulted in the highest accuracy. In the multi-view datasets, the clustering time is included in the total time, as the comparison is done with the methods that do not require clustering. In the single-view datasets, total time does not include the time used for clustering, as comparison is done to other clustering-based methods, where the same subclass labels are used. It can be seen that the proposed single-view method is performing better or close to the conventional methods, while always taking less time. 

\begin{table}[!h]
\caption{\label{tab3}Classification results of linear methods in single-view datasets: accuracy/number of clusters per class.}
\footnotesize\setlength{\tabcolsep}{5pt}
\centering
\begin{tabular}{|p{1.6cm}|p{0.5cm}p{0.1cm}|p{0.5cm}p{0.1cm}|p{0.5cm}p{0.1cm}|p{0.1cm}p{1cm}|p{0.5cm}p{0.1cm}|p{0.5cm}p{0.1cm}|}
\hline
\textbf{Dataset} & \multicolumn{2}{c|}{\textbf{SDA}} & \multicolumn{2}{c|}{\textbf{CDA}} & \multicolumn{2}{c|}{\textbf{SMFA}}  & \multicolumn{2}{c|}{\textbf{SRDA}} & \multicolumn{2}{c|}{\textbf{SDA,}} & \multicolumn{2}{c|}{\textbf{fastSDA}}\\
& &&&& & && & \multicolumn{2}{c|}{\textbf{sort. vec.}} & \multicolumn{2}{c|}{\textbf{(our)}}\\
\hline
\textbf{BU} & 62.8&1 & 60.1&1 & 59.9&1 && 62.6 & $\textbf{63.3}$&$\textbf{1}$ & $\textbf{63.3}$&$\textbf{1}$ \\
\hline
\textbf{Jaffe} & 65.2&1 & 58.1&1 & 63.8&1&  & 65.7 & 65.2&1 & $\textbf{66.2}$&$\textbf{1}$\\
\hline
\textbf{Ionosphere} & $\textbf{89.7}$&$\textbf{3}$ & 89.4&5 & 89.4&4 & &83.1 & 87.8&6 & 88.3&2 \\
\hline
\textbf{Kanade} & 63.3&1 & 61.6&1 & 55.1&1  && 65.3 & 64.0&1 & $\textbf{65.7}$&$\textbf{1}$\\
\hline
\textbf{Semeion} & 87.8&1 & 83.2&1 & 86.7&1  & & 88.9 & 89.0&1 & $\textbf{89.4}$&$\textbf{1}$\\
\hline
\textbf{Yale} & 86.8&2 & 86.6&2 & 87.6&2  && 88.6 & 88.7&1 & $\textbf{89.4}$&$\textbf{1}$\\
\hline
\textbf{PIMA} & 71.2&5 & 72.0&5& $\textbf{72.8}$&$\textbf{2}$  && 71.2 & 71.2&1 & 71.6&4\\
\hline
\textbf{Monks2} & 55.8&2 & 53.9&1 & $\textbf{61.2}$&$\textbf{1}$ & & 50.9 & 58.8&6 & 52.7&3\\
\hline
\textbf{SoF} & 98.6&1 & 98.9&1 & 98.5&1  && $\textbf{99.0}$ & 98.0&1 & $\textbf{99.0}$&$\textbf{1}$\\
\hline
\end{tabular}
\end{table}

\begin{table*}[!h]
\caption{\label{tab3t}Classification results of linear methods in single-view datasets: training time (in sec).}
\footnotesize\setlength{\tabcolsep}{5pt}
\centering
\begin{tabular}{|p{1.6cm}|P{1cm}|P{1cm}|P{1cm}|P{1cm}|P{1.5cm}|P{1.5cm}|}
\hline
\textbf{Dataset} & \textbf{SDA} & \textbf{CDA} & \textbf{SMFA}  & \textbf{SRDA} & \textbf{SDA, sort. vec.} & \textbf{fastSDA (our)}\\
\hline
\textbf{BU} & 0.019 & 0.017 & 0.030 & 0.013 & 0.09 & $\textbf{0.005}$ \\
\hline
\textbf{Jaffe} & 0.013 & 0.004 & 0.005  & 0.005 & 0.013 & $\textbf{0.002}$\\
\hline
\textbf{Ionosphere} & 0.008 & $\textbf{0.002}$ & 0.005 & 0.005 & 0.017 & $\textbf{0.002}$ \\
\hline
\textbf{Kanade} & 0.012 & 0.005 & 0.006  & 0.005 & 0.02 & $\textbf{0.002}$\\
\hline
\textbf{Semeion} & 0.045 & 0.041 & 0.147  & 0.015 &1.148 & $\textbf{0.013}$\\
\hline
\textbf{Yale} & 0.063 & 0.056 & 0.216  & 0.010 & 4.1 & $\textbf{0.007}$\\
\hline
\textbf{PIMA} & 0.003 & 0.009& 0.016 & 0.005 & 0.081 & $\textbf{0.001}$\\
\hline
\textbf{Monks2} & 0.004 & 0.002 & 0.002  & 0.005 & 0.005 & $\textbf{0.001}$\\
\hline
\textbf{SoF} & 9.52 & 18.3 & 86.0  & 0.831 & 0.801 & $\textbf{0.611}$\\
\hline
\end{tabular}
\end{table*}

\begin{table*}[!h] %no ! and no * for table in text. also remove 1 cm from each of the columns + 0.5 pt instead of 5 pt
\caption{\label{tab4}Classification results of kernel methods in single-view datasets: accuracy/number of clusters per class.}
\footnotesize\setlength{\tabcolsep}{5pt}
\centering
\begin{tabular}{|p{1.6cm}|p{0.15cm}p{0.5cm}p{0.6cm}|p{0.15cm}p{0.5cm}p{0.6cm}|p{0.15cm}p{0.5cm}p{0.6cm}|p{1cm}p{0.5cm}p{1.4cm}|}
\hline
\textbf{Dataset} & \multicolumn{3}{c|}{\textbf{kernel SDA}} & \multicolumn{3}{c|}{\textbf{kernel CDA}} & \multicolumn{3}{c|}{\textbf{kernel SMFA}} & \multicolumn{3}{c|}{\textbf{kernel fastSDA (our)}} \\
%& & & & & & & &\\
\hline
\textbf{BU} && 63.7&1 && $\textbf{64.7}$&${1}$ & &62.4&1 && 64.2&1\\
\hline
\textbf{Jaffe} && $\textbf{69.0}$&${1}$ && $\textbf{69.0}$&${1}$ & &63.8&1 & &68.5&1 \\
\hline
\textbf{Ionosphere} & &83.4&6 & &94.5&5 && 84.6&3 && $\textbf{94.9}$&${6}$ \\
\hline
\textbf{Kanade} && 59.6&2 & &60.4&1 &&  57.9&1 && $\textbf{61.2}$&${1}$ \\
\hline
\textbf{Semeion} && 91.2&2 && 91.5&1 && $\textbf{91.6}$&${1}$ && 90.6&1 \\
\hline
\textbf{Yale} && 89.4&6 && $\textbf{91.4}$&${1}$ && 75.2&4 && $\textbf{91.4}$&${1}$\\
\hline
\textbf{PIMA} && 63.1&6 && 66.9&6 && 64.8&5 && $\textbf{72.3}$&${3}$\\
\hline
\textbf{Monks2} && 46.0&6 && $\textbf{56.4}$&${5}$ && 55.2&2 && 52.7&3\\
\hline
\textbf{SoF} && 77.4&2 && 79.2&2 && 98.3&2 && $\textbf{98.4}$&${2}$\\
\hline
\end{tabular}
\end{table*}

\begin{table*}[!h] %no ! and no * for table in text. also remove 1 cm from each of the columns + 0.5 pt instead of 5 pt
\caption{\label{tab4t}Classification results of kernel methods in single-view datasets: training time (in sec).}
\footnotesize\setlength{\tabcolsep}{5pt}
\centering
\begin{tabular}{|p{1.6cm}|P{2cm}|P{2cm}|P{2cm}|P{3.5cm}|}
\hline
\textbf{Dataset} & \textbf{kernel SDA} & \textbf{kernel CDA} & \textbf{kernel SMFA} & \textbf{kernel fastSDA (our)} \\
\hline
\textbf{BU} & 0.036 & 0.068 & 0.432 &$\textbf{0.01}$\\
\hline
\textbf{Jaffe} & 0.004 &0.010 & 0.015 & $\textbf{0.001}$ \\
\hline
\textbf{Ionosphere} & 0.012 & 0.026 & 0.049 &$\textbf{0.002}$\\
\hline
\textbf{Kanade} & 0.005 & 0.007 &  0.020 & $\textbf{0.001}$ \\
\hline
\textbf{Semeion} & 0.275 & 0.479 & 11.5 & $\textbf{0.043}$ \\
\hline
\textbf{Yale} & 1.00 & 0.886 & 39.5 & $\textbf{0.103}$\\
\hline
\textbf{PIMA} & 0.040 & 0.392 & 0.368 & $\textbf{0.012}$\\
\hline
\textbf{Monks2} & 0.02 & 0.127 & 0.007 & $\textbf{0.001}$\\
\hline
\textbf{SoF} & 188.9 & 190.3 & 167.7 & $\textbf{1.55}$\\
\hline
\end{tabular}
\end{table*}

\begin{table*}[!h]
\caption{\label{tab5}Classification results of linear methods in multi-view datasets: accuracy/number of clusters per class.}
\centering
\footnotesize\setlength{\tabcolsep}{5pt}
\begin{tabular}{|p{2cm}|P{1.5cm}|P{1.5cm}|P{0.1cm}P{0.5cm}P{0.1cm}|P{0.9cm}P{0.5cm}P{0.1cm}|}
\hline
\textbf{Dataset} & {\textbf{SMvDA}} & {\textbf{MvMDA}} & \multicolumn{3}{c|}{\textbf{MvSDA (our)}} & \multicolumn{3}{c|}{\textbf{single-view fastSDA (our)}} \\
\hline
\textbf{HWD} & $\textbf{98.9}$ & 98.6 && 98.8&1 && 98.5&4\\
\hline
\textbf{HARS} & 62.6 & 31.9 && $\textbf{67.3}$&${1}$ && 63.0&3 \\
\hline
\textbf{Robots} & 66.8 & 57.5 && $\textbf{74.6}$&${5}$ && 46.4&6 \\
\hline
\textbf{Caltech-7} & $\textbf{98.2}$ & $\textbf{98.2}$ && $\textbf{98.2}$&${1}$ && 97.0&1 \\
\hline
\textbf{Caltech-20} & 93.7 & 94.6 && $\textbf{95.0}$&${1}$ && 89.7&1 \\
\hline
\textbf{HOPAR 1} & 84.9 & 84.8 && $\textbf{85.4}$&${1}$ && 84.8&2\\
\hline
\textbf{HOPAR 2} & 81.9 & 81.9 && $\textbf{82.3}$&${2}$ && $\textbf{82.3}$&${6}$\\
\hline
\textbf{MSDI} & 57.6 & 57.0 && $\textbf{58.4}$&${6}$ && 58.3&2\\
\hline
\textbf{NUS-WIDE} & 48.6 & 47.3 && $\textbf{56.0}$&${3}$ && 26.0&3 \\
\hline
\end{tabular}
\end{table*}

\begin{table*}[!h]
\caption{\label{tab5t}Classification results of linear methods in multi-view datasets: training time (in sec).}
\centering
\footnotesize\setlength{\tabcolsep}{5pt}
\begin{tabular}{|p{2cm}|P{1.5cm}|P{1.5cm}|P{2.1cm}|P{3.9cm}|}
\hline
\textbf{Dataset} & \textbf{SMvDA} & \textbf{MvMDA} & \textbf{MvSDA (our)} & \textbf{single-view fastSDA (our)} \\
\hline
\textbf{HWD} & 3.3 & 2.3 & 0.10 & $\textbf{0.03}$\\
\hline
\textbf{HARS} & 27.4 & 22.3 & 1.34 & $\textbf{0.22}$ \\
\hline
\textbf{Robots} & 0.029 & 0.028 & 0.01 & $\textbf{0.002}$ \\
\hline
\textbf{Caltech-7} & 21.5 & 22.4 & 1.35 & $\textbf{0.65}$ \\
\hline
\textbf{Caltech-20} & 23.4 & 20.2 & 2.0 & $\textbf{1.0}$ \\
\hline
\textbf{HOPAR 1} & 7.14 & 6.2 & 0.04 & $\textbf{0.008}$\\
\hline
\textbf{HOPAR 2} & 5.75 & 4.41 & 0.06 & $\textbf{0.008}$\\
\hline
\textbf{MSDI} & 58.4 & 50.9 & 0.2 & $\textbf{0.024}$\\
\hline
\textbf{NUS-WIDE} & 133.2 & 130.2 & 0.54 & $\textbf{0.09}$ \\
\hline
\end{tabular}
\end{table*}

\begin{table*}[!h]
\caption{\label{tab6}Classification results of kernel methods in multi-view datasets: accuracy/number of clusters per class.}
\centering
\footnotesize\setlength{\tabcolsep}{5pt}
\begin{tabular}{|p{2cm}|P{2.5cm}|P{2.5cm}|P{0.5cm}P{0.5cm}P{0.1cm}|P{0.9cm}P{0.5cm}p{1.2cm}|}
\hline
\textbf{Dataset} & {\textbf{kernel SMvDA}} & {\textbf{kernel MvMDA}} & \multicolumn{3}{c|}{\textbf{kernel MvSDA (our)}} & \multicolumn{3}{c|}{\textbf{kernel fastSDA (our)}} \\
\hline
\textbf{HWD} & 99.0 & 98.5 && $\textbf{99.3}$&${1}$ && 99.0&3\\
\hline
\textbf{HARS} & 79.4 & 86.5 && $\textbf{89.5}$&${3}$ && $\textbf{89.5}$&${2}$ \\
\hline
\textbf{Robots} & 68.3 & 75.2 && $\textbf{81.5}$&${2}$ && 77.6&4\\
\hline
\textbf{Caltech-7} & 97.6 & $\textbf{97.9}$  && 97.7&1 && 97.8&1\\
\hline
\textbf{Caltech-20} & 87.2 & 93.6 && 93.9&1 && $\textbf{94.7}$&${1}$ \\
\hline
\textbf{HOPAR 1} & 85.4 & $\textbf{86.0}$ && $\textbf{86.0}$&${2}$  && 85.8&2 \\
\hline
\textbf{HOPAR 2} & $\textbf{83.1}$ & 79.0 && 80.2&4  && 82.7&3 \\
\hline
\textbf{MSDI} & 51.3 & 31.6 && 61.5&1  && $\textbf{63.9}$&${1}$ \\
\hline
\textbf{NUS-WIDE} & 32.9 & 42 && 61.3&1  && $\textbf{62.7}$&${1}$ \\
\hline
\end{tabular}
\end{table*}

\begin{table*}[!h]
\caption{\label{tab6t}Classification results of kernel methods in multi-view datasets: training time (in sec).}
\centering
\footnotesize\setlength{\tabcolsep}{5pt}
\begin{tabular}{|p{2cm}|P{2.5cm}|P{2.5cm}|P{3.2cm}|P{3.2cm}|}
\hline
\textbf{Dataset} & \textbf{kernel SMvDA} & \textbf{kernel MvMDA} & \textbf{kernel MvSDA (our)} & \textbf{kernel fastSDA (our)} \\
\hline
\textbf{HWD} & 72.4 & 70.5 & 5.7 & $\textbf{0.07}$\\
\hline
\textbf{HARS} & 561 & 554 & 97 & $\textbf{4.3}$ \\
\hline
\textbf{Robots} & 0.14 & 0.14 & 0.03 & $\textbf{0.001}$\\
\hline
\textbf{Caltech-7} & 30.9 & 30  & 2.78 & $\textbf{0.03}$\\
\hline
\textbf{Caltech-20} & 244 & 236 & 9.57 & $\textbf{0.11}$ \\
\hline
\textbf{HOPAR 1} & 76.7 & 74.4 & 10.9  & $\textbf{4.49}$ \\
\hline
\textbf{HOPAR 2} & 65.9 & 74.1 & 8.89  & $\textbf{2.8}$ \\
\hline
\textbf{MSDI} & 69.9 & 48.6 & $\textbf{1.16}$  & 2.3 \\
\hline
\textbf{NUS-WIDE} & 259.5 & 235.3 & 24  & $\textbf{7.7}$ \\
\hline
\end{tabular}
\end{table*}

\FloatBarrier

\begin{figure}[h!]
    \centering
        \includegraphics[width=\textwidth]{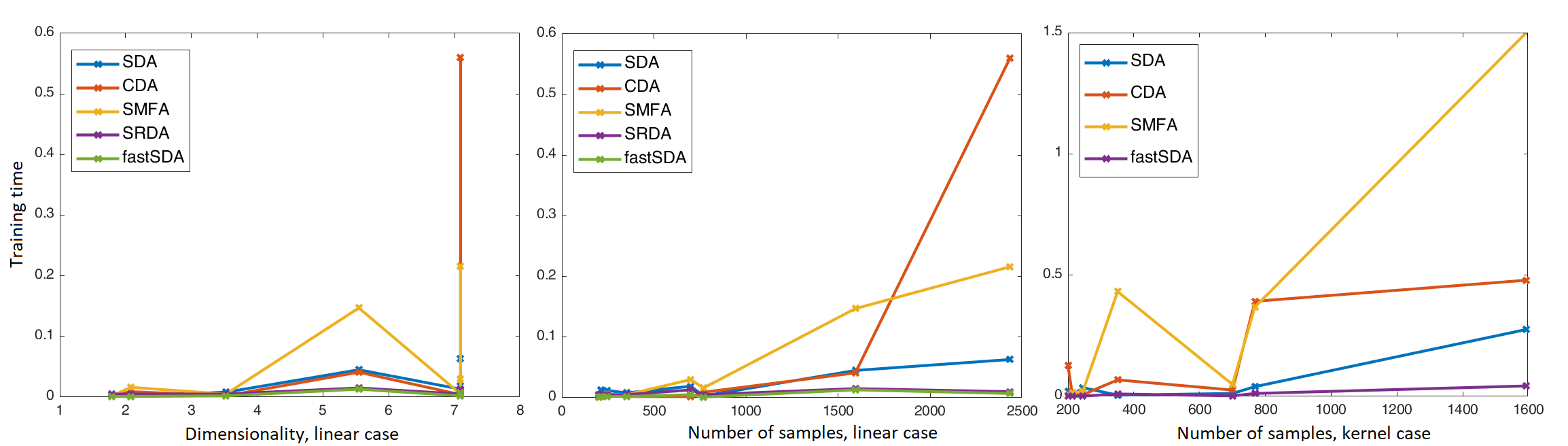}
        \caption{Dependency of training time on the dimensionality of data (left, dimensionality shown on log scale) and dependency of training time on the number of samples in the linear methods (middle) and kernel methods (right).}
        \label{lintimenumbersamples}
\end{figure}
\begin{figure}[h!]
    \centering
        \includegraphics[width=\textwidth]{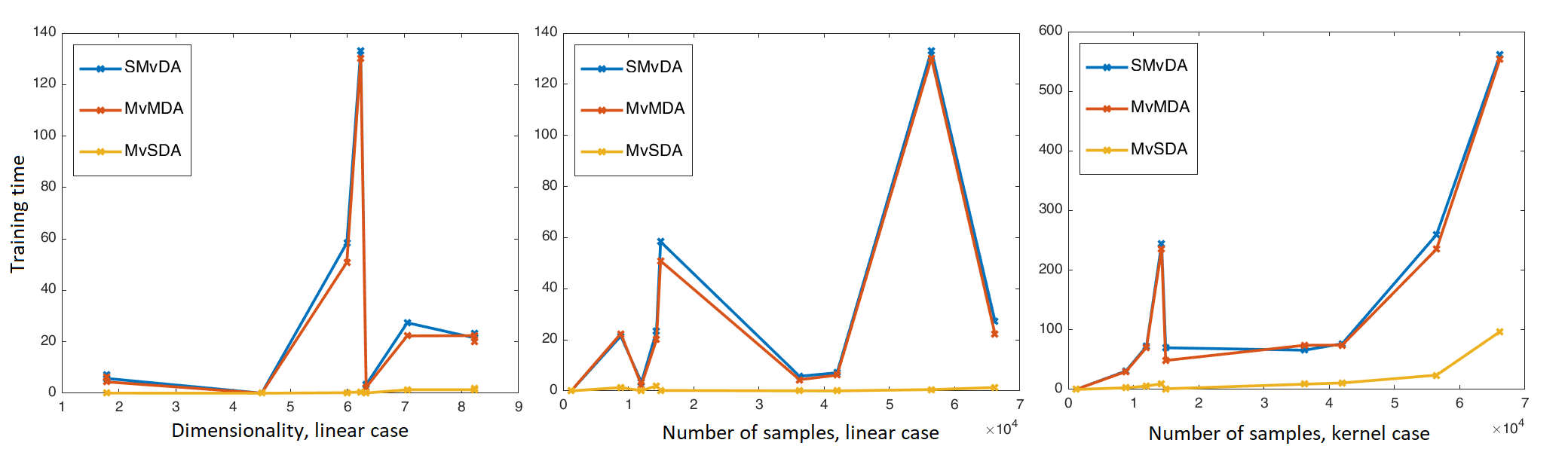}
        \caption{Dependency of training time on the dimensionality of data (left, dimensionality shown on log scale) and on the number of samples  (middle) in the linear multi-view methods and dependency of training time on the number of samples in the kernel multi-view methods (right). Numbers of samples and dimensionalities are summed across the views.}
        \label{lintimenumbersamples}
\end{figure}

Figures 9 and 10 show the dependency of the training time on the number of training samples in the dataset based on training on the datasets used in this work: Fig. 9 depicts the single-view methods, and Fig. 10 depicts the multi-view methods. Note that the speed of the methods is dependent on multiple factors, including the dimensionality, the number of samples, number of classes, and subclasses. Besides, in multi-view cases, the number of views in the dataset affects the training time significantly. Therefore, the training time is not always increasing gradually with increase in dimensionality/number of samples, as can be observed especially in the plots corresponding to linear formulations. However, it can be observed that the proposed methods outperform the existing approaches in terms of training time and the margin becomes higher with larger dataset sizes and dimensionalities. This can be seen especially well from the kernel formulation plots.

In the single-view linear case, the training time of fastSDA is similar to that of SRDA. However, accuracy-wise the proposed approach outperforms SRDA because of SRDA's assumptions on unimodality. This can be seen from Table 1. Otherwise, in the kernel formulation and in the multi-view scenarios, the proposed approach outperforms other methods in terms of computational complexity by a significant margin. 

In addition, by performing the projection onto the sorted by criterion value (11) regressed eigenvectors of $L_b$, we verify that for the data with subclass structure the eigenvectors corresponding to larger criterion values are those following the described structure. The only exceptions were observed in the Monks2 and PIMA datasets, where some of the eigenvectors had random structure - this is due to the samples of different subclasses being mixed with each other. However, even in this case, it can be observed that the proposed approach results in competitive  accuracy and higher speed. The accuracy obtained by projecting data using the transformation matrix comprised of eigenvectors corresponding to largest criterion values is shown in the second last column of Table~1.

For the multi-view case we compared the proposed multi-view SDA to other multi-view methods that assume unimodality of data. It can be seen that the proposed approach results in significant speed-up and competitive accuracy, often outperforming competing methods.
\section{Conclusions}
This work presents two contributions, proposing a fast and efficient solution for Subclass Discriminant Analysis and introducing multi-view Subclass Discriminant Analysis with a fast solution. As can be seen from the experimental results, the proposed speed-up approach allows to reduce the training time significantly, while being competitive in accuracy and often outperforming the conventional methods. Our findings allow performing the analysis on large-scale datasets, where conventional solutions are not feasible.
The proposed multi-view Subclass Discriminant Analysis provides superior accuracy compared to the methods relying on the assumption of unimodality of data. In addition, the proposed speed-up approach can be applied to this formulation, resulting in a significant gain in speed.

\section{Acknowledgement}
This work was done within the Center for Visual and Decision Informatics (CVDI) project Co-Botics jointly sponsored by Tieto Oy Finland and CA Technologies and the Business Finland project VIRPA-D sponsored by Tieto Oy Finland and other companies. Business Finland funding to NSF CVDI Project AMaLIA, Dnro 3333/31/2018 is gratefully acknowledged.

%\section{References}

\bibliography{bibliography}

\bibliographystyle{unsrt}

\end{document}